\documentclass{aifrontiers}
\usepackage{aifrontiers}
\usepackage{booktabs}
\usepackage{array}
\usepackage{tabularx}
\usepackage{wrapfig}
\usepackage{colortbl}
\usepackage{caption}
\usepackage{subcaption}
\usepackage{multirow}
\usepackage{fontawesome5}
\usepackage{float}
\usepackage{enumitem}
\usepackage{xspace}
\usepackage{soul}
\usepackage{listings}
\usepackage[export]{adjustbox}
\usepackage{algorithm}
\usepackage{algorithmic}
\usepackage{placeins}
\usepackage{needspace}
\usepackage{xurl}
\usepackage{titletoc}
\usepackage[T1]{fontenc}
\usepackage[utf8]{inputenc}
\usetikzlibrary{arrows.meta,positioning,calc}

\newcommand{\tabref}[1]{Table~\ref{#1}}
\newcommand{\figref}[1]{Figure~\ref{#1}}
\newcommand{\secref}[1]{Section~\ref{#1}}
\newcommand{\benchmark}{PersonaCodeBench}
\newcommand{\code}[1]{\texttt{#1}}
\newcommand{\met}{\mathsf{Satisfied}}

\newcolumntype{Y}{>{\raggedright\arraybackslash}X}
\newcolumntype{P}[1]{>{\raggedright\arraybackslash}p{#1}}
\definecolor{pipelineblue}{HTML}{315A83}
\definecolor{pipelinefill}{HTML}{EFF4F9}
\definecolor{csGreen}{HTML}{237567}
\definecolor{csBlue}{HTML}{315A83}
\definecolor{csGray}{HTML}{607386}
\definecolor{csAmber}{HTML}{936621}
\newcommand{\csUserTitle}[2]{%
  {\color{csGreen}\faUser\enspace User}\hfill
  {\normalfont\scriptsize\color{csGray}Turn #1\enspace\textbullet\enspace #2}}
\newcommand{\csAgentTitle}[1]{%
  {\color{csBlue}\faRobot\enspace coding agent}\hfill
  {\normalfont\scriptsize\color{csGray}#1}}
\newcommand{\csTurn}[2]{%
  \par\Needspace{10\baselineskip}\vspace{9pt}
  \noindent{\color{csGray!30}\rule[.5ex]{.22\linewidth}{.4pt}}\hfill
  \tcbox[on line,colback=csGray!7,colframe=csGray!7,boxrule=0pt,
    arc=3mm,left=8pt,right=8pt,top=3pt,bottom=3pt,boxsep=0pt]
    {\sffamily\footnotesize\color{csGray}\textbf{TURN #1}\enspace / \enspace #2}%
  \hfill{\color{csGray!30}\rule[.5ex]{.22\linewidth}{.4pt}}\par\smallskip}
\newcommand{\csUserAvatar}{%
  \fill[csGreen!6] ([yshift=-13pt]frame.north west) --
    ([xshift=-6pt,yshift=-18pt]frame.north west) --
    ([yshift=-23pt]frame.north west) -- cycle;
  \node[circle,fill=csGreen,text=white,inner sep=0pt,minimum size=20pt,
    font=\small] at ([xshift=-17pt,yshift=-17pt]frame.north west) {\faUser};}
\newcommand{\csAgentAvatar}{%
  \fill[csBlue!6] ([yshift=-13pt]frame.north east) --
    ([xshift=6pt,yshift=-18pt]frame.north east) --
    ([yshift=-23pt]frame.north east) -- cycle;
  \node[circle,fill=csBlue,text=white,inner sep=0pt,minimum size=20pt,
    font=\small] at ([xshift=17pt,yshift=-17pt]frame.north east) {\faRobot};}
\tcbset{
  csbase/.style={enhanced,breakable,boxrule=0pt,arc=2mm,
    left=9pt,right=9pt,top=6pt,bottom=6pt,
    before skip=7pt,after skip=7pt,
    coltitle=black,fonttitle=\small\sffamily\bfseries,
    fontupper=\small\setstretch{1.12}\raggedright,
    attach title to upper={\par\smallskip},
    pad at break*=2mm},
  csuser/.style={csbase,colback=csGreen!6,colframe=csGreen!6,arc=3mm,
    left skip=29pt,right skip=65pt,top=8pt,bottom=8pt,
    fontupper=\small\setstretch{1.12}\raggedright\setlength{\parskip}{3pt},
    overlay unbroken={\csUserAvatar},overlay first={\csUserAvatar},
    title after break={\color{csGreen}User\enspace{\normalfont\scriptsize(continued)}}},
  csagent/.style={csbase,colback=csBlue!6,colframe=csBlue!6,arc=3mm,
    left skip=65pt,right skip=29pt,top=8pt,bottom=8pt,
    fontupper=\small\setstretch{1.12}\raggedright\setlength{\parskip}{3pt},
    overlay unbroken={\csAgentAvatar},overlay first={\csAgentAvatar},
    title after break={\color{csBlue}coding agent\enspace{\normalfont\scriptsize(continued)}}},
  csinternal/.style={csbase,colback=white,colframe=csGray!30,
    borderline west={.6pt}{0pt}{csGray!30},arc=0pt,
    left skip=74pt,right skip=35pt,top=2pt,bottom=2pt,
    before skip=3pt,after skip=3pt,colupper=csGray,coltitle=csGray,
    fonttitle=\scriptsize\sffamily\bfseries,
    fontupper=\footnotesize\setstretch{1.05}\raggedright},
  csdecision/.style={csbase,colback=csAmber!5,colframe=csAmber!5,arc=1.5mm,
    left skip=65pt,right skip=29pt,top=4pt,bottom=4pt,
    before skip=3pt,after skip=7pt,colupper=csAmber,coltitle=csAmber,
    fonttitle=\scriptsize\sffamily\bfseries,
    fontupper=\footnotesize\setstretch{1.05}\raggedright}
}

\newcommand{\modellogo}[1]{%
  \makebox[1.45em][l]{\raisebox{-0.2em}{%
    \includegraphics[width=1.05em]{logo/providers/#1.pdf}}}}

\newcommand{\personaicon}[1]{{\color{pipelineblue}\faIcon{#1}}}
\newcommand{\personaname}[2]{%
  \begingroup\renewcommand{\arraystretch}{0.85}%
  \begin{tabular}[c]{@{}l@{}}%
    \textbf{#1}%
    \if\relax\detokenize{#2}\relax\else
      \\[-1pt]{\scriptsize\color{black!65}#2}%
    \fi
  \end{tabular}\endgroup}
\newcommand{\personascore}[3]{%
  \begingroup\renewcommand{\arraystretch}{0.85}%
  \begin{tabular}[c]{@{}c@{}}%
    #3\%\\[-1pt]{\scriptsize\mdseries\color{black!65}#1/#2}%
  \end{tabular}\endgroup}
\newcommand{\personanosamples}{{\color{black!45}\textendash}}

\begin{document}
\title{PersonaCodeBench: Evaluating Coding Agents in Persona-driven Human-Interactive Sessions}
\shorttitle{PersonaCodeBench}
\author{
\begin{minipage}{\textwidth}
\centering
\setlength{\parskip}{0pt}
{\fontsize{9.5}{12}\selectfont
\mbox{\textbf{Haixin Wang}\textsuperscript{1}}\hspace{0.6em}
\mbox{\textbf{Jianheng Hou}\textsuperscript{2}}\hspace{0.6em}
\mbox{\textbf{Hongyu Shen}\textsuperscript{3}}\hspace{0.6em}
\mbox{\textbf{Wenzhe Fan}\textsuperscript{4}}\hspace{0.6em}
\mbox{\textbf{Wei-Chieh Huang}\textsuperscript{4}}\hspace{0.6em}
\mbox{\textbf{Qianyu Zhu}\textsuperscript{5}}\hspace{0.6em}
\mbox{\textbf{Weichen Li}\textsuperscript{6}}\hspace{0.6em}
\mbox{\textbf{Taoran Li}\textsuperscript{7}}\hspace{0.6em}
\mbox{\textbf{Yuhang Yao}\textsuperscript{8}}\hspace{0.6em}
\mbox{\textbf{Ruohan Wang}\textsuperscript{9}}\hspace{0.6em}
\mbox{Zhejian Peng\textsuperscript{3}}\hspace{0.6em}
\mbox{Wanting Jiang\textsuperscript{10}}\hspace{0.6em}
\mbox{Yifan Chen\textsuperscript{11}}\hspace{0.6em}
\mbox{Shushu Liang\textsuperscript{12}}\hspace{0.6em}
\mbox{Haoming Li\textsuperscript{13}}\hspace{0.6em}
\mbox{Defu Cao\textsuperscript{2}}\hspace{0.6em}
\mbox{Xu Hu\textsuperscript{13}}\hspace{0.6em}
\mbox{Siyang Wu\textsuperscript{6}}\hspace{0.6em}
\mbox{Chen Yang\textsuperscript{12}}\hspace{0.6em}
\mbox{Haodi Hu\textsuperscript{2}}\hspace{0.6em}
\mbox{Haichao Zhang\textsuperscript{16}}\hspace{0.6em}
\mbox{Cong Wang\textsuperscript{12}}\hspace{0.6em}
\mbox{Zhiyuan Yu\textsuperscript{7}}\hspace{0.6em}
\mbox{Weizhi Zhang\textsuperscript{4}}\hspace{0.6em}
\mbox{Yifan Wang\textsuperscript{14}}\hspace{0.6em}
\mbox{Jintao Huang\textsuperscript{15}}\hspace{0.6em}
\mbox{Yizhou Sun\textsuperscript{1}}\hspace{0.6em}
\mbox{Yijun Wang\textsuperscript{12}}\hspace{0.6em}
\mbox{Yuexing Hao\textsuperscript{5,\textdagger}}\hspace{0.6em}
\mbox{Xiaomin Li\textsuperscript{12,\textdagger}}\par}
\vspace{6pt}
{\fontsize{8}{10}\selectfont
\mbox{\textsuperscript{1}University of California, Los Angeles}\hspace{0.7em}
\mbox{\textsuperscript{2}University of Southern California}\hspace{0.7em}
\mbox{\textsuperscript{3}University of Illinois at Urbana-Champaign}\hspace{0.7em}
\mbox{\textsuperscript{4}University of Illinois Chicago}\hspace{0.7em}
\mbox{\textsuperscript{5}Massachusetts Institute of Technology}\hspace{0.7em}
\mbox{\textsuperscript{6}University of Chicago}\hspace{0.7em}
\mbox{\textsuperscript{7}Texas A\&M University}\hspace{0.7em}
\mbox{\textsuperscript{8}Carnegie Mellon University}\hspace{0.7em}
\mbox{\textsuperscript{9}University of the Arts London}\hspace{0.7em}
\mbox{\textsuperscript{10}Amherst College}\hspace{0.7em}
\mbox{\textsuperscript{11}Northwestern University}\hspace{0.7em}
\mbox{\textsuperscript{12}Harvard University}\hspace{0.7em}
\mbox{\textsuperscript{13}The University of Texas at Dallas}\hspace{0.7em}
\mbox{\textsuperscript{14}Purdue University}\hspace{0.7em}
\mbox{\textsuperscript{15}The Ohio State University}\hspace{0.7em}
\mbox{\textsuperscript{16}Northeastern University}\par}
\vspace{4pt}
{\fontsize{8}{10}\selectfont\textbf{Bold}: Core Contributor\quad Contact: \href{mailto:whx@ucla.edu}{whx@ucla.edu}, \href{mailto:xiaominli@g.harvard.edu}{xiaominli@g.harvard.edu}\par}
\end{minipage}
}
\renewcommand{\AIFTitleAuthorGap}{1em}
\renewcommand{\AIFAbstractTopGap}{14pt}
\renewcommand{\weblink}{}
\renewcommand{\foundrylink}{}
\renewcommand{\hflink}{}
\renewcommand{\ghlink}{}

\begin{abstract}
Coding agents must deliver correct software while respecting users' expertise, preferences, and working conditions. Existing benchmarks increasingly incorporate interaction, yet personalized requirement fulfillment remains underexplored.
We introduce \benchmark{}, which turns user characteristics into verifiable obligations in persona-driven multi-turn coding sessions.
Our persona pool spans roles, technical abilities, interaction preferences, and practical constraints, yielding task-grounded \emph{evolved requirements} for implementation, permissions, testing, explanation, and delivery.
At runtime, a controller governs user simulator responses and requirement disclosure, using online requirement evaluation to provide feedback and advance the interaction. Requirements emerge progressively, with their content, disclosure conditions, and acceptance criteria fixed before evaluation.
A completion verifier and a personalization verifier separately assess native correctness and persona fulfillment. Joint pass requires both adherence to the declared budget and compliance with validity conditions.
We evaluate all major frontier agents, establish a comprehensive leaderboard, and conduct detailed experimental analyses to validate effectiveness.
\benchmark{} supports reproducible evaluation of coding agents' ability to meet explicit, personalized user expectations alongside coding objectives.
\par\vspace{4pt}
\begin{flushleft}
\small
\renewcommand{\arraystretch}{1.35}
\begin{tabular}{@{}c@{\hspace{0.6em}}l@{\hspace{0.8em}}l@{}}
\faGithub & \textbf{GitHub} &
\href{https://github.com/MatrAIx-ai/PersonaCodeBench}{\nolinkurl{https://github.com/MatrAIx-ai/PersonaCodeBench}} \\
\end{tabular}
\end{flushleft}
\end{abstract}
\maketitle

\begin{figure}[H]
\centering
\includegraphics[width=\textwidth]{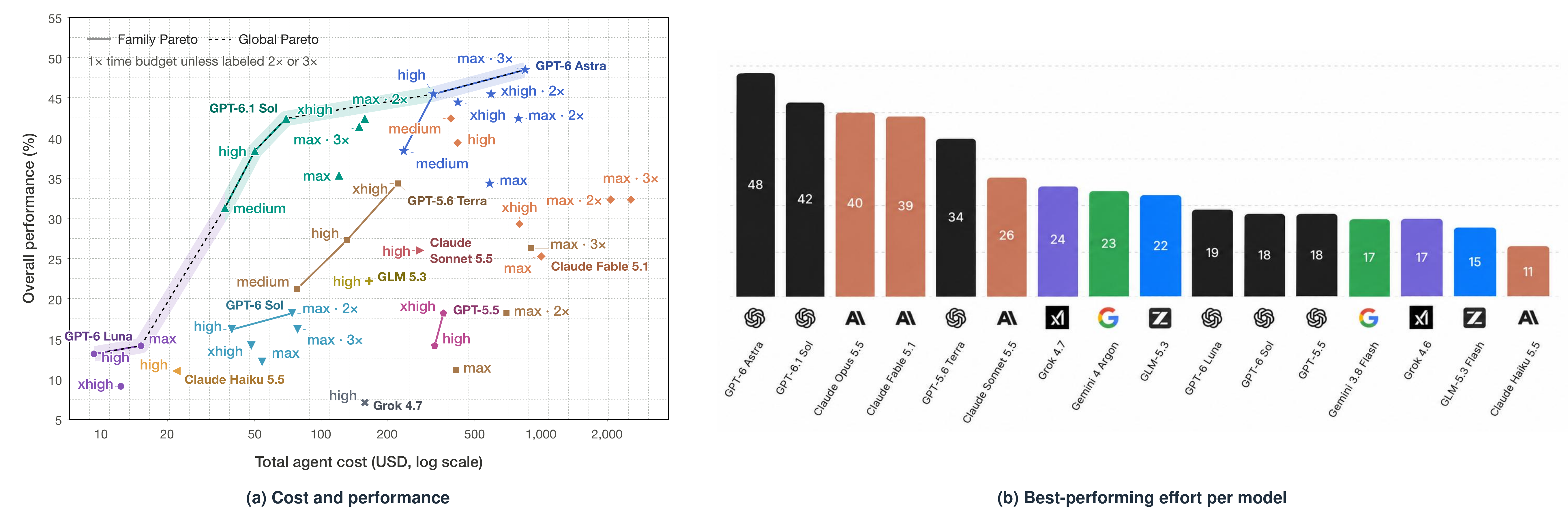}
\caption{\benchmark{} results overview. \textbf{(a)} Agent cost versus Joint pass rate, with reasoning effort and time-budget multipliers annotated. \textbf{(b)} Joint pass rates at the best-performing effort shown for each model.}
\label{fig:results-overview}
\label{fig:pareto}
\end{figure}

\section{Introduction}
\label{sec:intro}

Real coding sessions involve users clarifying goals, inspecting changes, and requesting revisions~\citep{baumann2026swechat}. A correct patch can remain unsuitable for its user: an analyst may lack permission to change a schema, a student may need an explanation, and a shop owner may be unable to install software. Success therefore depends on completing the task in a way that respects the user's expertise, preferences, and working conditions.

SWE-bench and its successors measure executable task completion~\citep{jimenez2024swebench,deng2025swepro,zhang2025swebenchlive}. SWE-Together and SWE-Interact introduce reactive feedback and progressive disclosure, with SWE-Interact also modeling a user persona~\citep{wu2026swetogether,raghavendra2026sweinteract}. Copilot Arena collects developer preferences in actual workflows~\citep{chi2025copilotarena}. Building on these advances, we ask: \emph{can a coding agent complete the software task while satisfying the particular user's requirements?} Controlled evaluation must connect user characteristics to observable obligations and distinguish functional correctness from personalized acceptance.

We introduce \benchmark{}, which turns user characteristics into verifiable coding constraints. Our self-constructed persona pool combines ten shared coding scenarios with structured user profiles. Profiles capture roles, expertise, and working constraints, together with recorded interaction, trust, explanation, and testing preferences. Stable attributes guide communication, while compatible preferences generate task-specific \emph{evolved requirements}. For a first-time contributor, these can require isolating the patch from existing changes, explaining the repair, and accurately reporting validation. Each obligation fixes its task bindings, disclosure trigger, dependencies, and acceptance evidence before evaluation.

Our construction pipeline pairs source tasks and personas, grounds added requirements in the task's environment, and simplifies the opening where appropriate while retaining the native specification for disclosure. Review checks compatibility, feasibility, and agreement between disclosed obligations and their checkers, repairing or rejecting unsuitable bindings. Contrastive screening targets tasks that a reference agent solves natively but fails under persona adaptation. We retain one binding per original task, yielding 231 pairs from five benchmarks, and select a source-proportional Lite release of 100 pairs with 186 added requirements.

During evaluation, the controller governs disclosure and authorization and validates proposed user messages. The user simulator expresses permitted requests, and the online requirement evaluator assesses the agent's response. Unmet obligations elicit feedback. Satisfied progression conditions permit the next eligible requirement. The final completion verifier measures native correctness, while the personalization verifier measures persona fulfillment. Joint pass requires both task completion and acceptance of all mandatory evolved requirements within the declared budget and validity conditions. Personalization is an operational proxy for explicit user expectations, and human satisfaction requires separate validation.

We evaluate mainstream proprietary and open-weight models, reporting a leaderboard and cost--performance comparisons. The highest reported Joint pass rate is 48\% on Lite. Composition analyses, disclosure ablations, and trajectory and reliability audits examine the agent behavior and evaluation process behind these scores.

Our contributions are fourfold:
\begin{itemize}[leftmargin=1.5em,topsep=4pt,itemsep=3pt]
  \item \textbf{A coding benchmark centered on user experience.} We introduce \benchmark{} and a reproducible construction process turning persona characteristics into verifiable constraints in multi-turn coding sessions.
  \item \textbf{Separate measures of completion and personalization.} We independently measure native correctness and persona fulfillment, requiring both for Joint pass under explicit evidence and budget policies.
  \item \textbf{Comprehensive model evaluation and a leaderboard.} We benchmark mainstream models across source subsets and reasoning settings, reporting rankings and cost--performance tradeoffs.
  \item \textbf{Analysis of construction and evaluation reliability.} We analyze task, persona, and requirement coverage, compare disclosure designs, and audit simulator delivery, checkers, and judge consistency through trajectory analyses and blinded AI-assisted review.
\end{itemize}

\begin{figure}[t]
\centering
\includegraphics[width=0.97\textwidth]{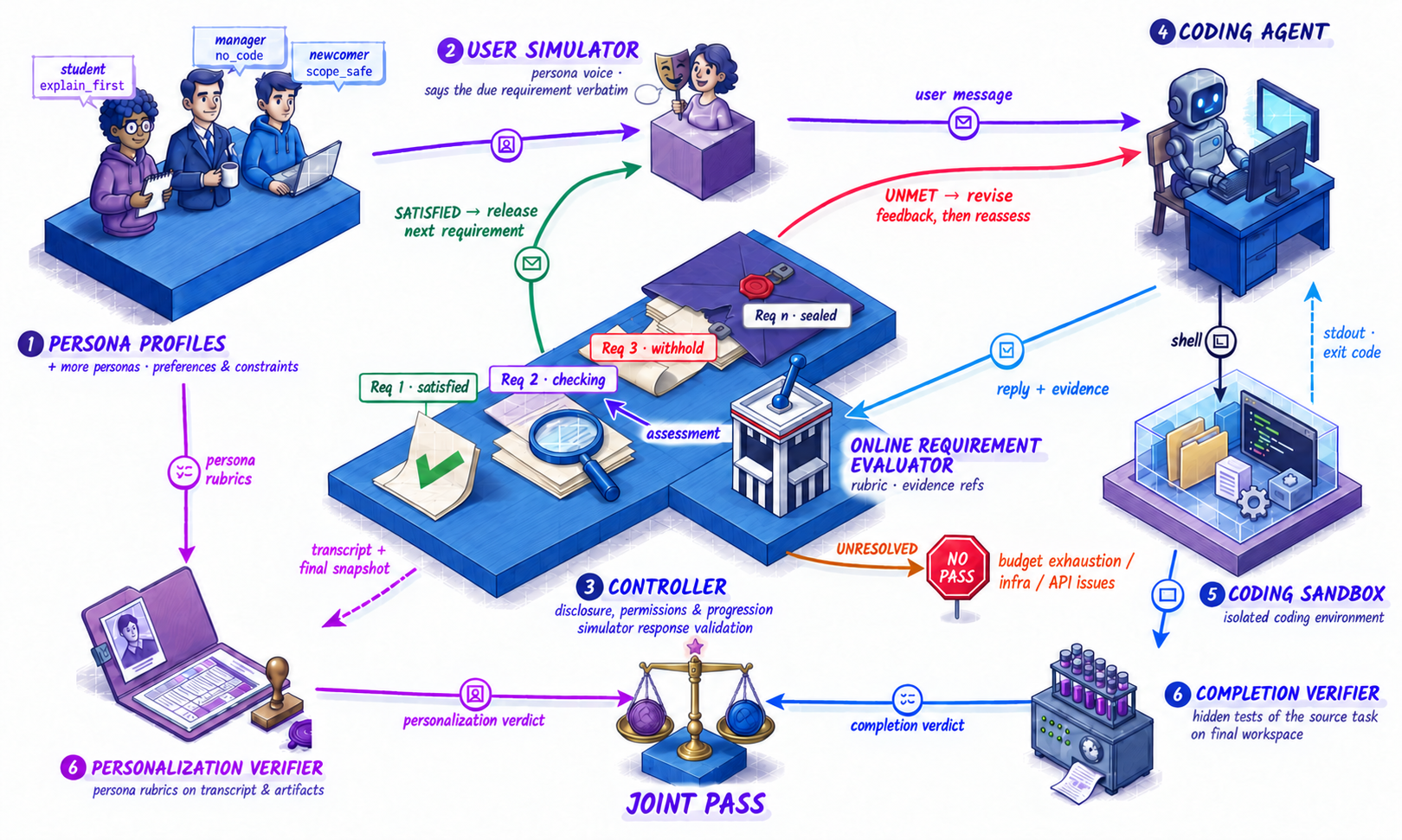}
\vspace{-0.3cm}
\caption{Overview of the persona-driven multi-turn coding sessions. The controller governs the disclosure of evolved requirements and uses the online requirement evaluator to validate responses. Satisfied requirements permit advancement and Unmet requirements return feedback to the coding agent for revision.}
\label{fig:framework}
\vspace{-0.3cm}
\end{figure}

\section{PersonaCodeBench}
\label{sec:method}

\subsection{Motivation}
\label{sec:motivation}

Coding-agent evaluation is expanding from single-turn implementation to multi-turn collaboration. SWE-Together reconstructs real user--agent sessions and evaluates both final correctness and corrective feedback~\citep{wu2026swetogether}. SWE-Interact uses a persona-conditioned simulator that begins with incomplete instructions, inspects the workspace, and progressively reveals requirements~\citep{raghavendra2026sweinteract}. These benchmarks establish interaction as an important evaluation dimension. A complementary question concerns personalized acceptance: \textbf{\textit{Can an agent complete a task in a way that satisfies the particular user's needs?}} 
Thus, we extend persona modeling along two key dimensions: (1) persona diversity, by introducing a broader spectrum of coding-related user personas, and (2) persona dynamics, by modeling each persona's task-specific expectations as evolving over time and explicitly incorporating these evolved requirements into the acceptance criteria.

In practical coding sessions, an initial request may describe the desired outcome without specifying every constraint on how it should be achieved or delivered. Those constraints can become explicit as the user reviews the agent's work. A student may ask for an explanation after receiving a patch, and a test-focused lead may request regression evidence before accepting it, and a shop owner who cannot install software may need a delivery path compatible with that restriction. 
Such \emph{evolved requirements} connect the user's role, expertise, and working conditions to concrete obligations. A functionally correct implementation can therefore remain unsuitable for its intended user. Modeling this process requires persona information to influence what the agent must accomplish, as well as how the simulator communicates.


\begin{figure}[!t]
\centering
\includegraphics[width=\textwidth]{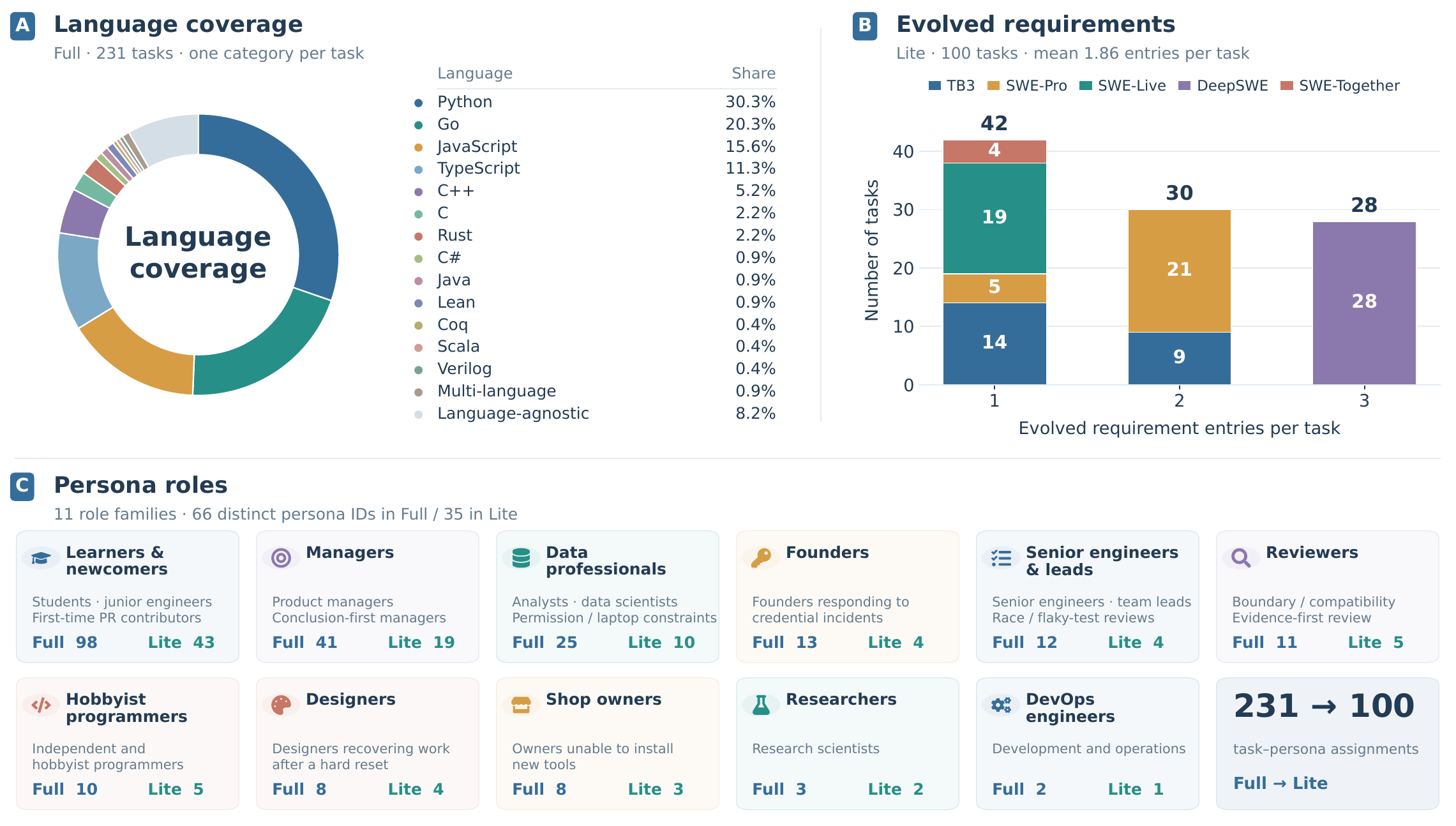}
\caption{Metadata of \benchmark{}. (A) language coverage. (B) evolved requirement counts. (C) persona role families.}
\label{fig:dataset-statistics}
\vspace{-0.3cm}
\end{figure}

\subsection{Persona Construction}
\label{sec:persona-construction}

\paragraph{Personas as specifications for collaboration.}
We define a persona through the user's role, technical background, goals, behavioral preferences, and working constraints. These attributes describe what the user can assess independently, how they prefer to collaborate, and what makes a solution useful in their setting. Personality-related descriptors, such as sociability, emotional tone, and conscientiousness, shape communication where they are explicitly recorded. Expertise, permissions, and evidence preferences additionally shape the work expected from the agent. A persona therefore influences both the interaction style and the obligations introduced during the session; it is not merely a static character description appended to the simulator prompt.

\paragraph{Constructing the persona pool.}
The shared scenario pool contains ten templates: a first-time PR contributor, a terse lead investigating flaky tests, a data scientist working on a laptop, a student requiring explanations, an outcome-oriented manager, an analyst without permission to create database indexes, a designer recovering from a destructive Git reset, a founder responding to a leaked production credential, a shop owner unable to install software, and a senior engineer requesting a concurrency review without introductory tutorials. These templates cover differences in technical independence, available resources, operational authority, and expected deliverables. The Terminal-bench 3 construction additionally uses structured profiles covering professional background, personality, coding preferences, and tool-specific behavior, including researchers, hobbyist programmers, and DevOps engineers. The shared templates are part of a broader pool, whose role families are summarized in \figref{fig:dataset-statistics}.

\paragraph{Behavioral dimensions.}
We characterize the recorded profiles along five complementary dimensions. \textit{Interaction} describes preferences for step-by-step guidance, evidence or constraints first, concise exchanges, planning before execution, autonomous execution, or iterative dialogue. \textit{Expertise and background} distinguish users needing code guidance, users with business or design knowledge, explicit technical proficiency, domain/tool familiarity, advanced review skills, and the ability to inspect source evidence. \textit{Trust and skepticism} capture explicitly stated trust or doubts about unsupported claims. \textit{Explanation} distinguishes learning-oriented walkthroughs, operational evidence, concise outcomes, decision and uncertainty summaries, and requests to avoid basic tutorials. \textit{Verification} describes expectations for tests, usable outcomes, explanatory self-checks, specialist or recovery evidence, cost-aware checks, and review materials. These dimensions accommodate differences among users with the same role. They are descriptive categories rather than a common psychometric scale: unstated attributes remain unspecified, and task obligations are grounded in explicit preferences and constraints rather than inferred from demographic identity or occupation alone. Appendix~\ref{app:persona-decomposition} provides detailed definitions, annotation conventions, and coverage analysis and  \figref{fig:persona-decomposition} in that appendix compares their distributions. 

\paragraph{From profiles to evolved requirements.}
For each task--persona binding, we translate relevant attributes into concrete evolved requirements concerning the implementation, workflow, explanation, verification, or handoff. Each requirement specifies its applicability, task-specific bindings, disclosure trigger, acceptance conditions, required evidence, and feedback when Unmet. For example, a first-time contributor may initially request help fixing an issue, then ask the agent to isolate its patch from existing workspace changes, explain a concrete failure and its repair, and prepare a reviewable PR draft with accurate validation results. The persona thus shapes what becomes necessary as the coding session progresses. Requirements and their acceptance meanings are fixed before evaluation, while their disclosure and activation evolve according to the agent's progress and the prescribed interaction policy. In requirement-gated profiles, feedback from the online requirement evaluator supports revision of the current obligation before progression to the next eligible request. This separation preserves a reproducible evaluation target while exposing the agent to dynamically unfolding, personalized demands. Only attributes instantiated as explicit task obligations contribute to personalization acceptance; stylistic descriptions do not automatically become hidden scoring conditions.

\subsection{Task Construction}
\label{sec:construction}

\paragraph{Single-turn coding tasks.}
We represent a native single-turn task as $\mathcal{T}=(I,W_0,V)$, where $I$ is the task specification, $W_0$ is the initial execution environment, and $V$ is the verifier for the native task. 
The agent receives $I$ at the outset, works in $W_0$ under a declared budget, and submits a final workspace $w$ for verification. 
Here, single-turn refers to the absence of subsequent user feedback: the agent may still make many tool calls, edit files, and run tests. Success depends on the original task's acceptance criteria. 

\paragraph{From single-turn tasks to persona-driven sessions.}
We extend each native coding task $\mathcal{T}$ with a persona $p=(s,R)$, where $s$ denotes a static user profile describing the user's background, preferences, and constraints, and $R$ represents a set of task-specific evolved requirements.
The resulting persona-driven task is: $\mathcal{X}=(\mathcal{T},p)$.
Unlike the original single-turn setting, the agent interacts with a persona-driven user simulator across multiple turns.
While the static profile $s$ guides the user's behavior throughout the session, requirements in $R$ are progressively disclosed as the interaction unfolds, shaping the agent's workflow and expected deliverables.
These requirements and their acceptance criteria are predefined, ensuring that the interaction evolves without changing the underlying evaluation contract.

\begin{figure}[!h]
\centering
\includegraphics[width=\textwidth]{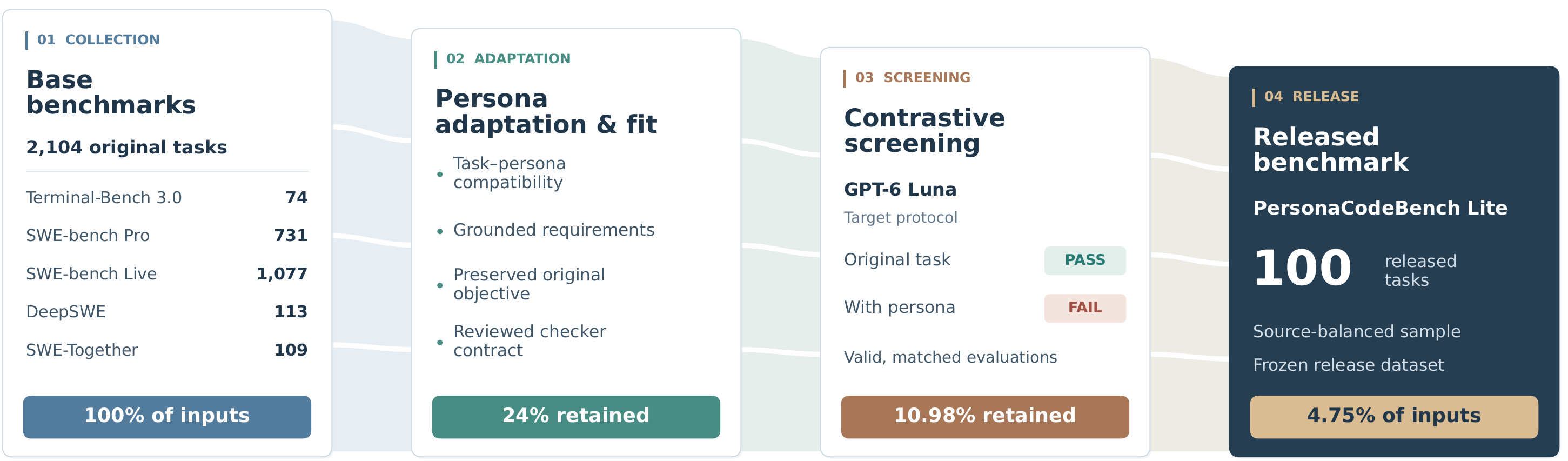}
\vspace{-0.2cm}
\caption{Schematic of the benchmark construction process.}
\label{fig:construction-pipeline}
\vspace{-0.3cm}
\end{figure}

\Needspace{6\baselineskip}

\paragraph{Screening and retention.} 
Building on the persona-driven, multi-turn agent–user interaction framework described above, we can systematically adapt existing coding tasks into persona-driven interactive sessions. \figref{fig:construction-pipeline} organizes construction into four stages: \emph{collection}, \emph{persona adaptation and fit}, \emph{contrastive screening}, and \emph{release}. Collection establishes the candidate pool from the source benchmarks. The subsequent stages address distinct questions: whether a task admits a meaningful, verifiable persona adaptation, and whether personalized interaction introduces a challenge beyond solving the native task.

At the adaptation \& fit stage, we review task--persona compatibility, the grounding of evolved requirements, preservation of the original objective, and agreement between the disclosed requests and the acceptance contract. 
Requirements must be feasible in the task environment and add observable obligations beyond the native specification. 
Unsupported persona demands, conflicting requirements, disclosure--acceptance mismatches, and insufficient verification evidence lead to rejection or repair. 
A compatible binding should admit a solution satisfying both the native and persona contracts:
\begin{equation}
  \exists(w,h):\quad V_T(w)=1\ \land\ \bigwedge_{r\in R} C_r(w,h)=\met,
  \label{eq:compatibility}
\end{equation}
where $h$ is the interaction history and $C_r$ checks requirement $r$. Native and jointly compliant references support feasibility checks, while targeted negative variants and semantic controls probe whether the corresponding evaluator rejects violated obligations and unsupported claims.

At the contrastive screening stage, the target protocol uses GPT-6 Luna as the reference agent and seeks the contrast \emph{native task passes, persona-adapted task fails}. The comparison requires valid, matched evaluations with the same original task and native acceptance criteria, under declared harness and budget settings. Failure of the adapted episode may reflect Unmet evolved requirements or loss of native correctness during personalized interaction.

\begin{wraptable}{r}{0.45\textwidth}
\centering
\scriptsize
\caption{Composition of the development collection and public Lite release.}
\label{tab:dataset}
\begin{tabular}{lrr}
\toprule
\textbf{Source benchmark} & \textbf{Full} & \textbf{Lite} \\
\midrule
Terminal-bench 3 & 52 & 23 \\
DeepSWE & 64 & 28 \\
SWE-bench-Live & 45 & 19 \\
SWE-Together & 9 & 4 \\
SWE-bench Pro & 61 & 26 \\
\midrule
\textbf{Total pairs} & \textbf{231} & \textbf{100} \\
evolved requirements & 422 & 186 \\
Distinct persona ID strings & 66 & 35 \\
\bottomrule
\end{tabular}
\end{wraptable}
\paragraph{Final collection.}
The integrated collection contains 231 distinct native tasks from 5 different sources, each paired with one frozen persona. We retain one persona binding per original task so that repeated adaptations do not duplicate a software problem in the evaluation set.
This sampling preserves the source proportions of the curated collection, rather than introducing another model-performance filter.

Full contains 422 evolved requirements, while Lite contains 186, and 42 Lite tasks have one requirement, 30 have two, and 28 have three. The collections contain 66 and 35 distinct persona ID strings, respectively. These identify concrete profiles rather than mutually exclusive behavioral categories. \figref{fig:dataset-statistics} summarizes language coverage, requirement counts, and persona role families. Proportional sampling does not remove the upstream selection effects or imply that the release represents all tasks in the source benchmarks.

\subsection{Persona-driven Agent--User Interaction}
\label{sec:simulator}

\figref{fig:framework} provides an overview of the interaction among the coding agent, user simulator, and controller. This section explains how the controller uses online requirement evaluation to coordinate requirement disclosure, feedback, and progression across coding turns. For the detailed execution flow and evidence boundaries, see \figref{fig:eval-flow} in Appendix~\ref{app:eval-flow}.

\paragraph{Controller-governed interaction.}
\benchmark{} separates conversational generation from interaction control and evaluation. The user simulator generates persona-consistent requests and feedback, while the controller maintains the interaction state, including disclosed requirements, active obligations, authorization, dependencies, and remaining budget. 
All task requirements and acceptance criteria are frozen before execution. The simulator may express these requirements in natural language but cannot independently introduce new grading conditions, grant unauthorized permissions, or advance the interaction.

\paragraph{Online requirement evaluation.}
After the coding agent acts, the host records its responses, tool executions, workspace changes, and relevant artifacts. 
At profile-defined checkpoints, the \textit{online requirement evaluator} assesses the currently active requirements against their frozen acceptance contracts using admissible evidence.
Depending on the requirement, evaluation uses deterministic rubrics for executable conditions or LLM judges for evidence-grounded semantic criteria. The evidence available to each evaluator is profile-specific: most profiles primarily assess eligible agent-visible responses, while SWE-Together can additionally inspect workspace diffs, artifacts, and execution records.

Each assessment returns a requirement verdict and, where applicable, remaining Unmet conditions. Conceptually, an obligation may be \underline{\textit{Satisfied}}, \underline{\textit{Unmet}}, or \underline{\textit{Unresolved}} due to insufficient evidence or unavailable evaluation. Exact verdict representations vary across profiles. Importantly, an online Satisfied verdict indicates progress on the current obligation, not final task completion or personalization acceptance.

The controller combines assessment outcomes with the frozen disclosure policy, requirement dependencies, authorization state, and remaining budget to determine the next interaction step. 
When an active obligation remains \underline{\textit{Unmet}}, the simulator can provide targeted feedback identifying the missing evidence or required corrections, allowing the agent to revise its work.
When the applicable progression conditions are \underline{\textit{Satisfied}}, the controller may authorize the disclosure of the next eligible requirement to the user simulator. This iterative assess--feedback--advance process ensures that user-specific demands are gradually introduced as the agent fulfills preceding requirements, rather than being disclosed all at once.
Besides, an \underline{\textit{Unresolved}} judgment follows the profile's specified retry, fallback, continuation, or termination policy rather than automatically counting as an agent failure. In such cases, further investigation is needed to identify the underlying cause of the failure. Common failure types include infrastructure issues, API timeouts, and other execution-related errors.

\begin{table*}[t]
\centering
\setlength{\tabcolsep}{2pt}
\renewcommand{\arraystretch}{1.12}
\caption{The pass rate pooled over all 100 tasks in \benchmark{}. \textsuperscript{\textdagger} only counts the textual part.}
\fontsize{7}{8.4}\selectfont
\label{tab:benchmark_comparison}
\begin{tabular*}{\textwidth}{@{\extracolsep{\fill}}lcc|*{5}{c}@{}}
\toprule
\textbf{Method}
& \textbf{Effort}
& \textbf{Joint pass}
& \textbf{Terminal-bench 3}
& \textbf{SWE-bench-Live}
& \textbf{DeepSWE}
& \textbf{SWE-bench Pro}
& \textbf{SWE-Together} \\
\midrule
\modellogo{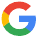}Gemini 4 Argon~\citep{google2026gemini4argon} & medium & 19.0 & 8.7 & 15.8 & 10.7 & 34.6 & 50.0 \\
\modellogo{google}Gemini 4 Argon~\citep{google2026gemini4argon} & high & 23.0 & 13.0 & 26.3 & 14.3 & 34.6 & 50.0 \\
\modellogo{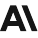}Claude Haiku 5.5~\citep{anthropic2026haiku55} & high & 11.0 & 0.0 & 36.8 & 0.0 & 11.5 & 25.0 \\
\modellogo{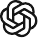}GPT-6.1 Sol~\citep{openai2026sol61} & medium & 31.0 & 4.3 & 52.6 & 14.3 & 53.8 & 50.0 \\
\modellogo{openai}GPT-6.1 Sol~\citep{openai2026sol61} & high & 38.0 & 17.4 & 47.4 & 39.3 & 42.3 & 75.0 \\
\modellogo{openai}GPT-6.1 Sol~\citep{openai2026sol61} & xhigh & 42.0 & 17.4 & 31.6 & 57.1 & 57.7 & 25.0 \\
\modellogo{anthropic}Claude Sonnet 5.5~\citep{anthropic2026sonnet55} & high & 26.0 & 8.7 & 26.3 & 0.0 & 65.4 & 50.0 \\
\modellogo{anthropic}Claude Opus 5.5~\citep{anthropic2026opus55} & medium & 40.0 & 8.7 & 57.9 & 21.4 & 53.8 & 50.0 \\
\modellogo{anthropic}Claude Opus 5.5~\citep{anthropic2026opus55} & high & 39.0 & 4.3 & 36.8 & 35.7 & 53.8 & 50.0 \\
\modellogo{openai}GPT-6 Sol~\citep{openai2026solandluna} & max & 18.0 & 0.0 & 31.6 & 0.0 & 46.2 & 0.0 \\
\modellogo{openai}GPT-6 Luna~\citep{openai2026solandluna} & medium & 19.0 & 0.0 & 31.6 & 7.1 & 34.6 & 50.0 \\
\modellogo{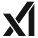}Grok 4.7~\citep{xai2026grok47} & high & 24.0 & 8.7 & 26.3 & 25.0 & 30.8 & 50.0 \\
\modellogo{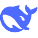}DeepSeek V4.1~\citep{deepseek2026v41} & max & 5.0 & 0.0 & 0.0 & 0.0 & 19.2 & 0.0 \\
\modellogo{openai}GPT-6 Astra~\citep{openai2026astra} & medium & 38.0 & 17.4 & 42.1 & 32.1 & 57.7 & 50.0 \\
\modellogo{openai}GPT-6 Astra~\citep{openai2026astra} & high & 45.0 & 17.4& 52.6 & 53.6 & 53.8 & 50.0 \\
\modellogo{openai}GPT-6 Astra~\citep{openai2026astra} & xhigh & 45.0 & 8.7 & 47.4 & 42.9 & 73.1 & 75.0 \\
\modellogo{openai}GPT-6 Astra~\citep{openai2026astra} & max & 48.0 & 17.4 & 36.8 & 53.6 & 80.8 & 25.0 \\
\modellogo{google}Gemini 3.8 Flash~\citep{google2026gemini38} & medium & 17.0 & 0.0 & 36.8 & 0.0 & 30.8 & 50.0 \\
\modellogo{google}Gemini 3.8 Flash~\citep{google2026gemini38} & high & 17.0 & 0.0 & 36.8 & 0.0 & 34.6 & 25.0 \\
\modellogo{anthropic}Claude Fable 5.1~\citep{anthropic2026fable51} & high & 39.0 & 4.3 & 36.8 & 39.3 & 69.2 & 50.0 \\
\modellogo{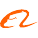}Qwen 3.8 Flash Next~\citep{alibaba2026qwen38flash} & medium & 1.0 & 0.0 & 0.0 & 0.0 & 3.8 & 0.0 \\
\modellogo{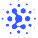}GLM-5.3 Flash~\citep{zai2026glm53flash} & max & 15.0 & 0.0 & 0.0 & 0.0 & 50.0 & 50.0 \\
\modellogo{alibaba}Qwen 3.8 27B~\citep{qwen2026qwen3827b} & medium & 6.0 & 0.0 & 5.3 & 0.0 & 15.4 & 25.0 \\
\modellogo{zhipu}GLM-5.3\textsuperscript{\textdagger}~\citep{zai2026glm53} & high & 22.0 & 4.3 & 31.6 & 0.0 & 50.0 & 50.0 \\
\modellogo{xai}Grok 4.6~\citep{xai2026grok46} & medium & 17.0 & 0.0 & 0.0 & 3.6 & 53.8 & 50.0 \\
\modellogo{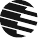}Kimi-K3~\citep{moonshot2026kimik3} & mixed & 6.0 & 0.0 & 0.0 & 0.0 & 15.4 & 50.0 \\
\modellogo{openai}GPT-5.6 Terra~\citep{openai2026terra56} & xhigh & 34.0 & 8.7 & 36.8 & 35.7 & 50.0 & 50.0 \\
\modellogo{openai}GPT-5.5~\citep{openai2026gpt55} & xhigh & 18.0 & 4.3 & 10.5 & 21.4 & 26.9 & 50.0 \\
\bottomrule
\end{tabular*}
\vspace{-0.2cm}
\end{table*}

\section{Experiments}
\label{sec:experiments}

\subsection{Evaluation setup.}
The evaluation uses the Lite subset. Each model receives the same task--persona assignments, source-specific scaffold, simulator/grading configuration, and declared budget policy. Exact deployment identifiers, decoding settings, repetitions, hardware, and budgets must be frozen before measurement. The public API changes the coding model through its agent configuration but retains each source's agent scaffold. We use GPT-6 Luna with medium reasoning effort for the user simulator and the judge model.

Our primary success criterion is Joint pass: the completion verifier must accept the original software task outcome, and the personalization verifier must accept all mandatory evolved requirements under the task's acceptance contract.
A successful episode must therefore both complete the task and satisfy the user's explicit needs, subject to the declared budget and evaluation-validity conditions. Passing only one verifier receives no joint credit. 
We use Harbor 0.23.0 and Docker environments~\citep{harbor2026}. Timing policies must be reported with the runtime revision: the later revision documented in Appendix~\ref{app:eval-flow} uses separate preparation, interaction, and verification watchdogs, with calibration and input materialization outside those clocks and source-specific inner limits still active.

\subsection{Main Results}
\label{sec:mainresult}

Table~\ref{tab:benchmark_comparison} reports the percentage of Lite tasks receiving a Joint pass. Under this reporting convention, every outcome other than a recorded pass contributes zero, including evaluation errors and unexecuted tasks. GPT-6 Astra at max effort achieves the highest reported Joint pass rate, 48\%, followed by GPT-6.1 Sol at xhigh (42\%) and Claude Opus 5.5 at medium (40\%). Claude Fable 5.1 and Opus 5.5 at high both reach 39\%. Even the strongest evaluated configuration passes fewer than half of the task--persona pairs. These results leave substantial headroom for joint task completion and personalization acceptance.

We also notice that the highest reported Terminal-bench 3 score is only 17.4\%, corresponding to four of its 23 tasks, whereas the best scores on SWE-bench Pro, SWE-bench-Live, and DeepSWE are 80.8\%, 57.9\%, and 57.1\%, respectively. Different configurations lead these sources: GPT-6 Astra/max on SWE-bench Pro, Claude Opus 5.5/medium on SWE-bench-Live, and GPT-6.1 Sol/xhigh on DeepSWE. Strong performance on one source therefore does not imply uniformly strong performance across the benchmark. For example, Claude Sonnet 5.5/high passes 65.4\% of SWE-bench Pro tasks but none of DeepSWE, while GPT-6.1 Sol/xhigh achieves 57.1\% on DeepSWE but 31.6\% on SWE-bench-Live. The pooled result weights these source-specific strengths by task count: DeepSWE and SWE-bench Pro together account for 54 of the 100 tasks. SWE-Together has only four tasks, so each success changes its score by 25 percentage points; its results support only a coarse comparison.

Qwen 3.8 Flash Next, Qwen 3.8 27B, and Kimi-K3 receive 1\%, 6\%, and 6\% overall, respectively, under the same fixed-denominator counting rule. 
Kimi additionally mixes original settings and high-effort repairs. These low recorded-pass yields, therefore combine task performance with evaluation coverage and should not be interpreted as failure rates on fully graded coding episodes alone.

\subsection{Persona Analysis}
\label{sec:persona-analysis}

\begin{wraptable}{r}{0.57\textwidth}
\centering
\scriptsize
\caption{Joint pass outcomes by persona and source subset for GPT-6 Luna.}
\label{tab:persona_subset_results}
\begingroup
\fontsize{7.5}{8.5}\selectfont
\setlength{\tabcolsep}{2pt}
\renewcommand{\arraystretch}{1.08}
\renewcommand{\tabularxcolumn}[1]{m{#1}}
\renewcommand{\personaname}[2]{\parbox[c]{80pt}{\raggedright\textbf{#1}\if\relax\detokenize{#2}\relax\else\par{\scriptsize #2}\fi}}
\begin{tabularx}{\linewidth}{@{}>{\centering\arraybackslash}m{11pt}
  @{\hspace{4pt}}>{\raggedright\arraybackslash}m{80pt}
  *{5}{>{\centering\arraybackslash}X}@{}}
\toprule
\multicolumn{2}{l}{\textbf{Persona}}
& {\fontsize{6.5}{8}\selectfont\textbf{\shortstack{Terminal-\\bench 3}}}
& {\fontsize{7}{8}\selectfont\textbf{\shortstack{SWE-\\bench Pro}}}
& \textbf{\shortstack{SWE-\\bench-\\Live}}
& \textbf{\shortstack{SWE-\\Together}}
& {\fontsize{6.5}{8}\selectfont\textbf{DeepSWE}} \\
\midrule
\rowcolor{pipelinefill}
\personaicon{graduation-cap} & \personaname{Student}{Requiring explanations}
& \personascore{0}{4}{0.0} & \personascore{1}{5}{20.0} & \personascore{0}{5}{0.0} & \personascore{0}{1}{0.0} & \personascore{1}{5}{20.0} \\
\addlinespace[1pt]
\personaicon{code-branch} & \personaname{First PR contributor}{}
& \personascore{0}{1}{0.0} & \personascore{3}{9}{33.3} & \personanosamples & \personascore{0}{1}{0.0} & \personascore{0}{8}{0.0} \\
\addlinespace[1pt]
\rowcolor{pipelinefill}
\personaicon{bullseye} & \personaname{Outcome-oriented manager}{}
& \personascore{0}{3}{0.0} & \personanosamples & \personascore{1}{2}{50.0} & \personanosamples & \personascore{1}{13}{7.7} \\
\addlinespace[1pt]
\personaicon{tasks} & \personaname{Test-focused lead}{}
& \personascore{0}{2}{0.0} & \personascore{2}{3}{66.7} & \personascore{0}{1}{0.0} & \personanosamples & \personanosamples \\
\addlinespace[1pt]
\rowcolor{pipelinefill}
\personaicon{search} & \personaname{Senior reviewer}{}
& \personascore{0}{2}{0.0} & \personanosamples & \personascore{1}{2}{50.0} & \personascore{2}{2}{100.0} & \personanosamples \\
\addlinespace[1pt]
\personaicon{database} & \personaname{Analyst}{Without index permissions}
& \personanosamples & \personascore{1}{4}{25.0} & \personascore{2}{2}{100.0} & \personanosamples & \personanosamples \\
\addlinespace[1pt]
\rowcolor{pipelinefill}
\personaicon{palette} & \personaname{Designer}{}
& \personascore{0}{6}{0.0} & \personascore{0}{2}{0.0} & \personascore{0}{1}{0.0} & \personanosamples & \personascore{0}{1}{0.0} \\
\addlinespace[1pt]
\personaicon{store} & \personaname{Shop owner}{Unable to install software}
& \personascore{0}{2}{0.0} & \personascore{0}{1}{0.0} & \personascore{0}{1}{0.0} & \personanosamples & \personascore{0}{1}{0.0} \\
\addlinespace[1pt]
\rowcolor{pipelinefill}
\personaicon{key} & \personaname{Founder}{With leaked credentials}
& \personanosamples & \personascore{2}{2}{100.0} & \personascore{1}{2}{50.0} & \personanosamples & \personanosamples \\
\addlinespace[1pt]
\personaicon{chart-line} & \personaname{Data scientist}{}
& \personascore{0}{3}{0.0} & \personanosamples & \personascore{1}{3}{33.3} & \personanosamples & \personanosamples \\
\midrule
\multicolumn{2}{l}{\textbf{Total}}
& \textbf{\personascore{0}{23}{0.0}}
& \textbf{\personascore{9}{26}{34.6}}
& \textbf{\personascore{6}{19}{31.6}}
& \textbf{\personascore{2}{4}{50.0}}
& \textbf{\personascore{2}{28}{7.1}} \\
\bottomrule
\end{tabularx}
\endgroup
\end{wraptable}

\paragraph{Separating persona difficulty from task difficulty.}
Table~\ref{tab:persona_subset_results} shows large differences in Joint pass rate: founders achieve 3/4, whereas students and first-time contributors achieve 2/20 and 3/19, and designers and shop owners achieve none. To distinguish personalization difficulty from task difficulty, we also inspect the component grades in the same GPT-6 Luna run. Among valid episodes with an observed persona grade, requirements pass for 4/4 founders, 4/4 test-focused leads, 5/6 analysts, and 4/5 senior reviewers, compared with 4/15 managers, 1/4 shop owners, 6/13 students, and 7/14 first-time contributors. These denominators exclude unavailable grades, unlike the assigned-task denominators in the table. The strongest evidence of difficulty concerns fulfilling operational and evidentiary obligations, rather than simply adopting a particular tone. The comparisons remain descriptive: tasks differ across roles, some reporting groups use an approximate crosswalk, and several groups are small.

\paragraph{More tractable demands: bounded, inspectable deliverables.}
Founders are the most successful group in this run, but their security-oriented framing does not imply that the agent resolved a production security incident. In the SWE-bench Pro founder episode \code{r0323}, the actual obligations are concrete: avoid introducing credentials or unauthorized endpoints, then provide a path-and-reason review list for every changed file. After one clarification about the list's exact format, both obligations pass final rechecking, as does the native task. The successful SWE-Together reviewer episode similarly proceeds from source inspection to explicitly authorized removal and then a commit, with inspected symbols, scope exclusions, and supporting tests stated in the handoff. These examples suggest why some demanding personas are still comparatively tractable: their requirements reuse evidence available from normal repository inspection and the final diff, and corrective feedback points to a bounded omission. High expertise or explicit skepticism does not necessarily make a user harder to satisfy.

\paragraph{Managers: concise reporting can hide missing acceptance evidence.}
The manager category has only 2/18 Joint pass outcomes and 4/15 observed personalization pass outcomes. The PSD blend-range trajectory exposes a failure beyond writing style: the native check passes, and the persona checks accept the three-point format and evidence-based progress update, but reject the final acceptance summary. The agent reports that probes passed without delivering the requested scenario-by-scenario user actions and visible outcomes. Thus, the information existed in execution records but was not translated into an actionable handoff. Concision must preserve the evidence a non-coding user needs to decide whether the result works. This pattern is particularly pronounced in DeepSWE, where only 1/11 observed manager persona grades passes; the source's staged acceptance requirements are therefore part of the explanation, not merely the role label.

\paragraph{First-time contributors: a correct patch is not a reviewable contribution.}
First-time contributors pass 6/6 observed persona checks in SWE-bench Pro but 0/7 in DeepSWE, despite sharing the reporting category. In DeepSWE's ABS module-cache episode, the native tests pass and the causal explanation of the fix is accepted. Nevertheless, the scope manifest does not correctly separate task changes from a pre-existing file, and the handoff omits earlier failed validation attempts even though it reports later passing checks. The failure is one of maintaining a complete contribution record across the interaction: final success does not erase earlier evidence, and a plausible explanation cannot replace an accurate scope inventory. This illustrates why requirements tied to execution history and changing artifacts can be harder than producing a final answer in the requested style.

\paragraph{Shop owners: operational constraints create state-management failures.}
Shop owners have 0/5 Joint pass outcomes and 1/4 observed personalization pass outcomes. Their trajectories require usable delivery under restricted installation and recovery conditions. In SWE-bench Pro \code{r0211}, the agent supplies a patch, preview, and rollback scripts; the dialogue eventually receives positive final judgments. Yet after rollback verification, the objective checker finds no remaining task changes, and both patch-export and rollback requirements fail with \code{no\_task\_change}. Verifying reversibility and preserving the final deliverable have diverged. In DeepSWE's SQLite safe-import episode, the owner workflow and preview checks pass, but the rollback check fails because the agent rehearses restoration on a different file with its own backup, rather than the specified sandbox and named original. It also omits the required confirmation stopping point. These failures concern exact paths, state transitions, and authorization boundaries; a reassuring description of a similar workflow is insufficient.

\paragraph{Low Joint pass rate need not mean an unsatisfied persona.}
Designers provide the clearest counterexample: they have no Joint pass outcomes, yet pass 4/5 observed persona checks, while all nine valid native evaluations fail. Students also show mixed failure modes. In the DataClaw Windows-support episode, the agent supplies the requested error explanation, boundary comparison, and self-check, satisfying the persona, but the native reward is only 0.8. These results do not support treating either group as uniformly difficult to satisfy. Across the inspected cases, the more persistent personalization failures arise when the agent must preserve execution history, maintain a valid final workspace, or translate tests into user-verifiable acceptance evidence. Those demands extend across turns and artifacts, whereas a bounded review list or source-grounded explanation can often be repaired through a focused clarification.

\subsection{Coding agent \& user simulator behavior analysis}
\label{sec:behavior}

We inspect selected public GPT-6.1 Sol/high trajectories to examine how simulator feedback changes the agent's subsequent behavior. 

\paragraph{Concrete feedback connects explanations to implementation evidence.}
In the SWE-Together DataClaw Windows-support episode, the agent initially describes several Windows-compatibility problems and proposes corresponding repairs. At user turn 2, the simulator asks for one failing path and its expected output before implementation. The agent then traces why a backslash-separated Windows home path misses a prefix constructed with forward slashes, exposing the username, and explains the intended project-relative result. 
When the next request asks about the changed source before any edit has occurred, the agent explicitly distinguishes its proposed correctness argument from an implemented fix. 
It also examines a neighboring directory name that must not match the Documents prefix and a Unix input whose behavior must be preserved. After implementation, the final walkthrough links the example and boundary cases to the actual source branches. Host execution records confirm 182 anonymizer/parser tests and 340 full-suite tests passing. The final native reward is 1.0 and personalization is accepted. The observed interaction therefore moves from a broad plan to a concrete, testable explanation while preserving the distinction between intended and completed work. It does not establish that a human learned from the explanation or that the feedback caused the functional success.

\paragraph{Evidence-view mismatches can create verification loops after a repair.}
In the SWE-Together ComfyUI Lumina2 LoRA-prefix episode, the agent first investigates the mapping behavior without editing, then receives explicit authorization to implement the missing alias. By agent turn 5, host records show the implementation and focused tests, including a transition from five failing tests to seven passing tests. The simulator nevertheless cites an empty \code{visible\_diff} observation and asks for fresh workspace evidence in user turns 5--7. Agent turns 6--8 consequently revisit the diff, untracked test file, checkout identity, and validation output. The three corresponding host inspection records have the same workspace snapshot hash, and the focused suite is rerun twice after its initial passing run. The agent eventually states that its shell evidence cannot establish the identity or provenance of the separate empty-diff observation, rather than asserting an unsupported explanation. Final native and personalization verification both pass. This case exposes an unresolved mismatch between observation channels; it does not identify which channel produced the mismatch. It also shows why interaction length needs contextual interpretation: additional turns can reconcile evidence rather than advance the implementation, and their cost can arise from the simulator's information view as well as the agent's behavior.

\subsection{Ablation Study}
\label{sec:ablation}

\begin{wraptable}{r}{0.59\textwidth}
\centering
\fontsize{8.5}{10}\selectfont
\setlength{\tabcolsep}{2pt}
\renewcommand{\arraystretch}{1.2}
\caption{Ablation Study of key interactive design on GPT-6 Luna.}
\label{tab:ablation}
\begin{tabular*}{\linewidth}{@{\extracolsep{\fill}}lccc@{}}
\toprule
Ablation & Joint pass & completion pass & \shortstack{personalization\\pass} \\
\midrule
\benchmark{} & 19.0\% & 24.0\% & 50.0\% \\
\shortstack[l]{\benchmark{}\\w. Opening} & 20.0\% & 26.0\% & 51.0\% \\
\shortstack[l]{\benchmark{}\\w/o Persona} & 45.0\% & 45.0\% & -- \\
\bottomrule
\end{tabular*}
\end{wraptable}

We compare the default \benchmark{} method with two ablated variants, using the default method as the baseline for both comparisons. \textbf{\benchmark{}} is the default method and uses the complete persona-driven interaction: requirements are disclosed progressively under the frozen policy, the agent must satisfy the current obligation before advancing through applicable gates, and final success requires both the completion verifier and personalization verifier to pass. \textbf{\benchmark{} w. Opening} gives the agent all evolved requirements in the first user message, in dependency order and with their applicable times. It retains the personalization acceptance criteria, online gates, and authorization requirements: knowing that approval will be needed does not authorize an action in advance. This variant examines whether earlier knowledge of the requirements improves performance. \textbf{\benchmark{} w/o Persona} removes the persona profile, persona-specific communication style, added obligations, and associated judges and grading. A neutral simulated user continues to provide the native task's clarification and feedback, and success is determined by native verification alone. This variant measures the combined effect of evolved requirements and their interaction machinery. It retains multi-turn native interaction.

\paragraph{Evolved requirements introduce substantial challenges.}
Table~\ref{tab:ablation} shows that removing evolved requirements raises the reported Joint pass rate from 19\% for \benchmark{} to 45\% for \benchmark{} w/o Persona, a gain of 26 percentage points. The completion pass rate also rises from 24\% to 45\%, indicating that the difficulty extends beyond an additional acceptance criterion: satisfying user-specific obligations while navigating the associated interaction process can also affect functional task completion. The benchmark therefore tests whether an agent can deliver a correct solution while meeting the user's constraints, rather than coding correctness alone. Because this ablation removes both the obligations and their associated gates and judges, the gain reflects the combined effect of that persona-driven workflow.

\paragraph{Progressive disclosure yields similar scores and models realistic interaction.}
Providing all evolved requirements at the opening changes the Joint pass rate only from 19\% to 20\%; completion pass and personalization pass rates likewise differ by two and one percentage points, respectively. In this single-run comparison, earlier access to the requirements offers little aggregate benefit, although it does not establish statistical equivalence between the two disclosure policies. We retain progressive disclosure because it models how users refine an initially incomplete request as they inspect an agent's plan, implementation, or proposed deliverable. Requirements become explicit at the relevant point in the conversation, while their acceptance criteria remain fixed in advance. This supports a more natural interaction scenario without making delayed disclosure the main observed source of difficulty.

\FloatBarrier
\section{Related Work}
\label{sec:related}

\subsection{Interactive Software Engineering}
Coding benchmarks differ in both the work they require and the feedback available to the agent. BigCodeBench introduces library-dependent tasks with complex instructions, while LiveCodeBench collects fresh contest problems and evaluates capabilities including repair and execution~\citep{zhuo2024bigcodebench,jain2024livecodebench}. SWE-bench moves evaluation to issue resolution in existing repositories~\citep{jimenez2024swebench}. SWE-bench Pro emphasizes substantial, long-horizon changes, and SWE-bench-Live broadens repository coverage through continuously collected issues and reproducible environments~\citep{deng2025swepro,zhang2025swebenchlive}. RepoLaunch addresses the associated build-and-test preparation bottleneck~\citep{li2026repolaunch}, while Terminal-bench and DeepSWE extend executable evaluation to demanding terminal workflows and original engineering tasks~\citep{terminalbench2026,deepswe2026}. These settings establish whether delivered software meets a specified objective; they do not by themselves separate that outcome from acceptance by users with different permissions, expertise, or handoff expectations.

Interactive coding also encompasses several distinct feedback mechanisms. SlopCodeBench evaluates successive extensions of an agent's own implementation under evolving specifications~\citep{orlanski2026slopcodebench}. ColBench introduces multi-turn collaboration on backend programming and frontend design, with a simulated collaborator informed by a reference artifact~\citep{zhou2025sweetrl}. SWE-Together reconstructs real coding sessions, conditions interventions on agent progress, and reports final correctness, corrective steering, and intent coverage~\citep{wu2026swetogether}. SWE-Interact explicitly models an Expert Nitpicker persona, inspects the workspace, and progressively introduces constraints; it also analyzes forgotten requirements, regressions, and rework, while retaining the original verifier as the task-success criterion~\citep{raghavendra2026sweinteract}. These works already establish persona-conditioned, dynamically unfolding collaboration. \benchmark{} adds task-grounded evolved requirements with frozen evidence and activation rules, and evaluates their acceptance jointly with native correctness. SWE-Together is itself one of our sources; our adaptations define a distinct evaluation protocol rather than an unchanged upstream score.

\subsection{Persona-driven User--agent Interaction}
Personalization evaluation spans user understanding, tailored outputs, and interactive preference discovery. LaMP studies personalized generation and classification using user histories, while PersonaBench tests recovery of personal information from synthetic private documents~\citep{salemi2024lamp,tan2025personabench}. PersonaLens evaluates task-oriented dialogues using profiles, preferences, and interaction histories, with separate user and judge agents assessing personalization, response quality, and task success~\citep{zhao2025personalens}. UserBench models underspecified travel goals and incrementally disclosed preferences, measuring whether agents elicit and use those preferences in tool-supported decisions~\citep{qian2025userbench}. Personalization and evolving user intent are therefore established evaluation targets, extending beyond static role descriptions.

Executable interaction benchmarks further distinguish successful actions from successful collaboration. $\tau$-bench verifies tool-mediated conversations against target database states, and $\tau^2$-Bench studies coordination when both the agent and user can act on a shared environment~\citep{yao2024taubench,barres2025tau2bench}. More recently, iPrefBench evaluates personalized mobile tasks whose missing requirements must either be inferred from user history or obtained by asking~\citep{ning2026iprefbench}. PersonaCUABench evaluates computer-use agents on fixed tasks through persona agents that reformulate requests and answer clarification questions, revealing gaps between task completion and persona-specific user satisfaction~\citep{li2026personacuabench}. Our focus is the coding-specific connection between a user's working conditions and verifiable deliverables: authorization before edits, isolation of existing changes, accurate validation reports, and explanations grounded in the actual patch. The persona determines which obligations apply, while the task binding specifies when they become active and what evidence establishes fulfillment. This complements preference retrieval and elicitation by testing whether personalized obligations remain compatible with, and are satisfied alongside, the original software objective.

\subsection{Online Evaluation of Human Satisfaction}
Evaluations with real users provide evidence that executable task scores alone cannot supply. Copilot Arena brings pairwise code-completion judgments into developers' IDEs and finds rankings that differ from offline evaluations; it also shows that latency affects whether users accept suggestions~\citep{chi2025copilotarena}. WildFeedback extracts preference signals from natural conversational feedback for alignment~\citep{shi2024wildfeedback}, and SWE-chat provides observational evidence about developer corrections and whether agent-written code survives into commits~\citep{baumann2026swechat}. These are different measurement settings: preference votes, naturally occurring feedback, and retained code each capture part of the user experience. Agent Arena extends online evaluation to live agent sessions, using randomized component assignments to estimate performance differences and separating explicit task confirmation, praise versus complaints, and steering effort. Its current steerability definition counts both correction-free turns and the burden of repeated corrections~\citep{arena2026agentarena,arena2026signals}.

Live feedback captures authentic expectations and deployment effects, whereas frozen task--persona contracts support repeated comparisons against explicit obligations. The two approaches provide complementary evidence. ACLArena studies how capabilities such as tool use and instruction following are retained or forgotten across stages of agent post-training~\citep{wang2026aclarena}. Its findings motivate evaluating explicit requirement fulfillment across model updates alongside direct measures of user experience. SimulatorArena directly examines whether simulated behavior and assistant ratings align with human conversations and judgments~\citep{dou2025simulatorarena}; Eval4Sim separately assesses persona adherence, consistency, and conversational naturalness against human dialogue~\citep{bao2026eval4sim}. Accordingly, \benchmark{} treats personalization acceptance as an operational measure of explicit requirement fulfillment, not an observed human-satisfaction rating. Disclosure controls and evidence-constrained grading make that measure inspectable, but agreement with real users requires independent validation. 

\section{Conclusion}
\label{sec:conclusion}

\benchmark{} makes user expectations explicit in executable coding evaluation. It binds personas to task-grounded requirements, controls their disclosure during interaction, and assesses native correctness and personalization acceptance through separate evidence contracts. The current resource combines 231 development pairs and a fixed 100-pair Lite release spanning five benchmarks and six execution profiles. The available statistics and release records characterize this resource, while \benchmark{} specifies how to compare task completion, personalization, and interaction effort. Completing that study, validating simulator and judge behavior with human review, and broadening task-level control coverage are the next steps toward understanding coding agents as collaborators under diverse user requirements.

\bibliographystyle{plainnat}
\bibliography{references}
\clearpage
\appendix

\startcontents[appendices]
\section*{Appendix}
\begingroup
\singlespacing
\printcontents[appendices]{}{1}{\setcounter{tocdepth}{2}}
\endgroup

\section{Detailed Evaluation Workflow}
\label{app:eval-flow}

\begin{figure}[H]
\centering
\includegraphics[width=\textwidth]{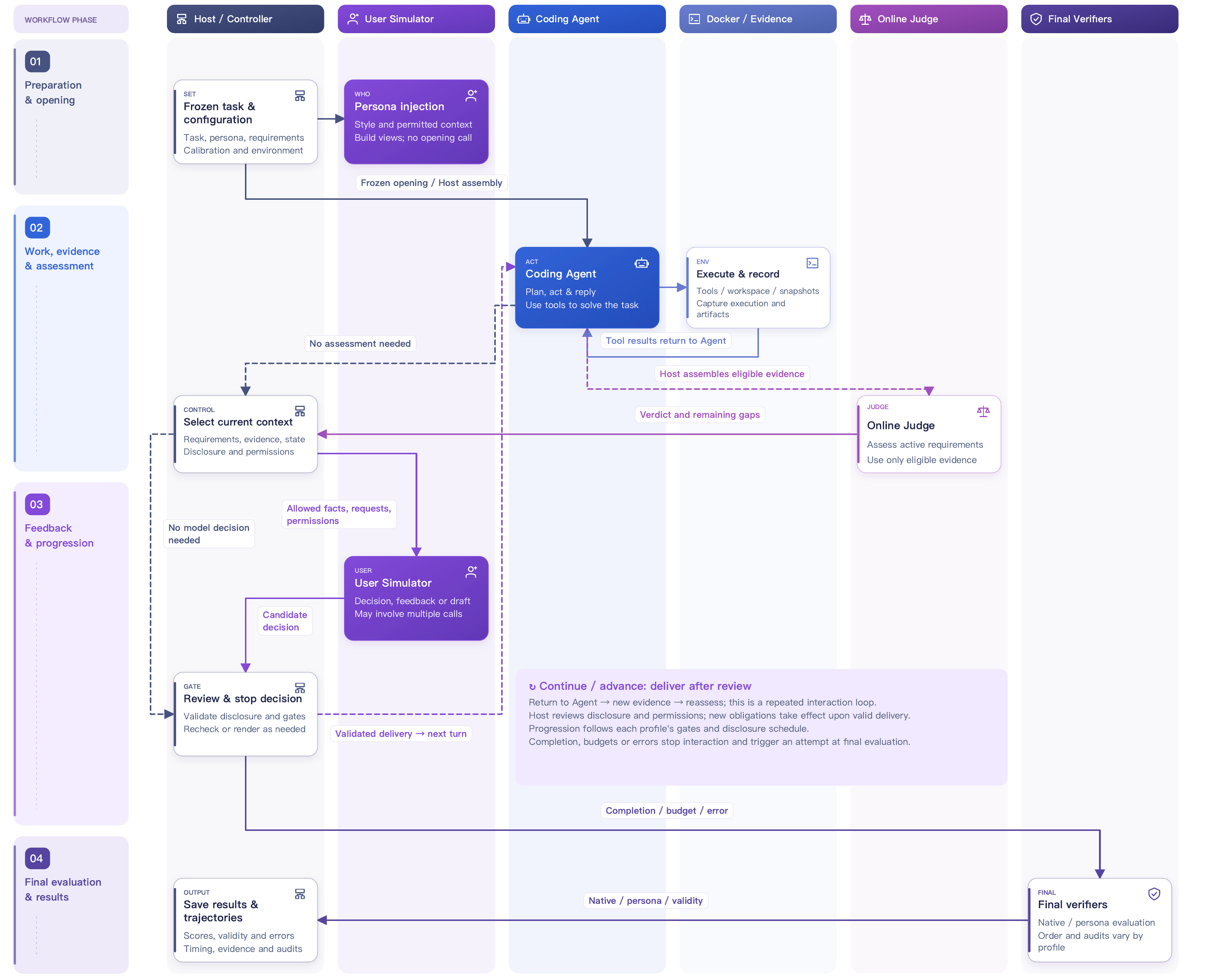}
\caption{Detailed evaluation flow across the six execution profiles. The online requirement evaluator assesses agent responses and the controller governs transitions and qingwen simulator responses. Final verifiers produce completion and personalization outcomes used to determine joint pass.}
\label{fig:eval-flow}
\end{figure}

To comprehensively demonstrate the multi-turn persona-driven evaluation, \figref{fig:eval-flow} traces a task--persona episode through four phases: (1) preparation and opening; (2) agent work, evidence collection, and online requirement evaluation; (3) feedback and controlled progression; and (4) final evaluation and result recording.
The flow documents control and evidence boundaries. It does not add end-to-end measurements or certify evaluator accuracy.

\subsection{Phase 1: Preparation and Opening}
\label{app:eval-preparation}

The unified entry selects a frozen task--persona pair, applies the profile's calibration admission policy, materializes its assets, and dispatches a formal worker. 
The binding fixes the native task, persona, evolved requirements, disclosure schedule, acceptance rules, environment, and budgets before interaction.
Runtime materialization restores these inputs. It does not generate new task-specific requirements. Calibration admission is profile-specific: permitted degraded admission and any provisional or independent-checker flags must remain visible in the run record. 

The worker prepares the appropriate Harbor/Docker environment and any required native-reference preflight or resource checks. The host constructs separate views for the agent, simulator, and evaluators. Possessing the complete binding does not authorize the host to expose future requests, hidden tests, or reference patches to those model inputs. Persona setup establishes the simulator's style and permitted context; constructing that context is not itself a model call.

The delivered opening is frozen text or a fixed host assembly. DeepSWE, Terminal-bench 3, both SWE-bench Pro profiles, and SWE-bench-Live reuse frozen openings or instructions. SWE-Together combines its native instruction with initial preference fields. SWE-bench-Live's applicability-fit call precedes interaction and does not regenerate the delivered opening. Thus, the start of a session must be distinguished from the first simulator-model call.

\subsection{Phase 2: Agent Work, Evidence, and online requirement evaluation}
\label{app:eval-evidence}

The coding agent plans, invokes tools, modifies permitted workspace state, and replies. Tool outputs return to the agent directly. A tool call need not create a user turn. The host records the transcript, commands and results, artifacts, and relevant workspace or revision identities. These records support different predicates: a completion claim is not an execution receipt, and evidence from an earlier revision may not establish the state of the delivered code.

At a profile-defined checkpoint, the host determines whether assessment is due and assembles only eligible evidence for active requirements. 
DeepSWE, Terminal-bench 3, SWE-bench Pro, and SWE-bench-Live primarily use eligible visible agent messages for online judgment, with disclosure and reference rules specific to each profile. SWE-Together can additionally use artifacts, diffs, and execution records. There is therefore no universal online-evidence bundle or verdict vocabulary.

The online requirement evaluator assesses the agent's response and eligible evidence and supplies a verdict and remaining gaps to the controller. A $\met$ verdict may permit progression, but does not establish native correctness or final personalization acceptance. Unavailable assessment follows a declared repair, fallback, continuation, or stop path. For example, Terminal-bench 3 can continue authorized work after an unavailable ordinary checkpoint, while SWE-bench-Live's deterministic fallback is restricted to its complete frozen handoff contract. Neither branch silently converts evaluator unavailability into a failed user obligation.

\subsection{Phase 3: Feedback and Controlled Progression}
\label{app:eval-progression}

The controller combines current evidence, requirement dependencies, disclosure timing, permissions, and remaining budget to select the next action. When language-model feedback is needed, the simulator receives the persona and permitted session view and proposes a response or decision. The controller reviews the proposed action against the profile's transition rules and performs simulator response validation before delivering the next user message. Response validation concerns the simulator's output; online requirement evaluation concerns the coding agent's output. Structural checks and any semantic LLM review remain profile-specific. Native task facts, evolved requirements, and implementation authorization remain separate states; a proposed request or permission becomes agent-visible only upon valid delivery.

The loop is conditional rather than a fixed evaluator--simulator sequence. SWE-Together may deliver due native follow-ups directly. SWE-bench Pro may accept a valid normal submission without a simulator call and executes host-approved explicit verification after review. DeepSWE separates a simulator action proposal, controller transition review, message rendering, and simulator response validation before delivery. A valid finish can skip rendering. Its S1--S4 labels are logical stages in a frozen plan, not four fixed conversation turns. Once feedback is delivered, the agent resumes work and produces new evidence for the next checkpoint.

\Needspace{8\baselineskip}
\subsection{Phase 4: Final Verification and Result Recording}
\label{app:eval-final}

Completion gates, budget exhaustion, or errors end interaction and trigger an attempt at final evaluation. The runtime preserves the stop reason and the available transcript and workspace evidence. An interruption need not leave a usable snapshot or permit all checks to finish. Normal simulator acceptance, online $\met$, and recovery eligibility are distinct conditions. Recovery uses the profile's own rules and is not automatically an ordinary successful completion.

The completion verifier evaluates the native task, and the personalization verifier evaluates the persona contract, independently of simulator approval. Deterministic checkers test executable predicates; LLM judges assess semantic obligations. Their outcomes determine Joint pass. Applicable integrity and delivery audits remain separate checks. \tabref{tab:eval-final-order} specifies the principal ordering differences: notably, SWE-Together performs persona finish and audit before freezing the container and running native tests. An online verdict may be reused only where the profile permits it and the evidence remains applicable to the final state.

\begin{table}[!htbp]
\centering
\small
\caption{Principal finalization order in the reviewed runtime. Arrows show the main path; conditional hooks and recovery branches remain profile-specific.}
\label{tab:eval-final-order}
\renewcommand{\arraystretch}{1.15}
\begin{tabularx}{\linewidth}{@{}P{2.9cm}Y@{}}
\toprule
\textbf{Profile} & \textbf{Finalization and verification} \\
\midrule
DeepSWE & Snapshot $\rightarrow$ completion verifier $\rightarrow$ personalization verifier. \\
SWE-Together & personalization verifier / audit $\rightarrow$ container freeze $\rightarrow$ completion verifier. \\
Terminal-bench 3 & Conditional hooks / frozen export $\rightarrow$ completion verifier $\rightarrow$ integrity gate $\rightarrow$ personalization verifier $\rightarrow$ integrity recheck. \\
SWE-bench Pro & Frozen export $\rightarrow$ completion verifier $\rightarrow$ personalization verifier. \\
SWE-bench-Live & Snapshot $\rightarrow$ completion verifier $\rightarrow$ personalization verifier (deterministic handoff checker on the last eligible reply). \\
\bottomrule
\end{tabularx}
\end{table}

The adapter and unified entry retain native and persona outcomes, validity, stop and error categories, evidence references, timings, and trajectories. Missing component grades remain unavailable. Audit, integrity, cleanup, and recovery results can affect the reported outcome, so adapter results must be retained alongside aggregate metrics. The reported Joint pass rate does not replace these profile-specific execution and verification rules.

\subsection{Timing and Outcome Boundaries}
\label{app:eval-timing}

The host publishes transitions among three separately supervised phases. Calibration and input materialization precede the formal worker and sit outside these watchdogs. The interaction cap is not a single timeout covering setup and all final grading; profile-specific coding, action, call, and turn limits also remain active.

\begin{itemize}[leftmargin=1.3em,topsep=3pt,itemsep=3pt]
  \item \textbf{Preparation.} Worker startup through the host's \code{begin\_interaction} transition, including environment preparation and applicable preflight. The default cap is 7,200 seconds; exhaustion is an infrastructure error with unavailable score.
  \item \textbf{Interaction.} From \code{begin\_interaction} to \code{begin\_verification}, including agent work, tools, simulator waits, and online requirement evaluation. The task budget supplies the default; \code{--timeout-sec} overrides the outer interaction cap. Exhaustion is a failed attempt, with any available component grades retained and missing grades left unavailable.
  \item \textbf{Verification.} Remaining final checks and audits after the host transition. The default cap is $\max(900,\,t_{\mathrm{verifier}}+600)$ seconds, where $t_{\mathrm{verifier}}$ is the configured verifier timeout. Exhaustion is an infrastructure error with incomplete verification, rather than an inferred native or persona failure.
\end{itemize}

The unified entry applies outer-timeout classification after adapter normalization, so a recovery path cannot override the watchdog outcome.

\section{Persona Decomposition and Attribute Coverage}
\label{app:persona-decomposition}

\begin{figure}[!htbp]
\centering
\includegraphics[width=\textwidth]{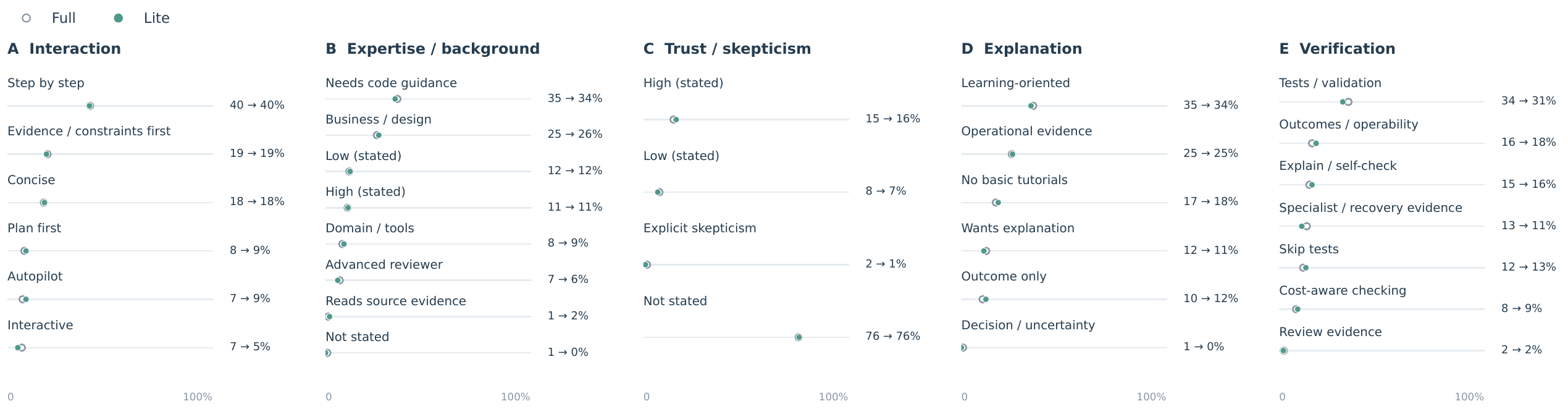}
\caption{Persona attribute composition in Full and Lite. Panels show interaction, expertise/background, trust/skepticism, explanation, and verification. Each assignment contributes one category per panel; annotations give rounded Full $\rightarrow$ Lite percentages.}
\label{fig:persona-decomposition}
\end{figure}

A persona combines expectations about how an agent should collaborate, explain its work, and support acceptance. A role name alone does not capture these expectations: users in similar roles can prefer different interaction styles or forms of verification. \figref{fig:persona-decomposition} summarizes the recorded profiles along five dimensions, complementing the task-contract formulation in \secref{sec:construction}.

\subsection{Representation and Counting}
\label{app:persona-counting}

The five dimensions provide a common descriptive view of heterogeneous source profiles. 
The aggregation maps recorded descriptors to panel categories, with broader verification groups combining related source codes such as patch validation, required tests, focused regression, and compatibility validation. 
Each task--persona assignment is counted once per dimension, including repeated assignments of the same persona ID. Categories summarize the recorded descriptors and do not exhaust everything a persona may express. Missing information remains \emph{Not stated}; it is not imputed from occupation or other preferences.

\subsection{Five Dimensions of Collaboration}
\label{app:persona-dimensions}

\paragraph{Interaction.}
Step-by-step interaction is the largest category in both collections (40\% after rounding), followed by evidence/constraints first (19\%) and concise interaction (18\%). Plan-first, autopilot, and interactive profiles capture additional preferences about sequencing and agent initiative. These distinctions motivate different communication patterns, but the frozen disclosure and authorization policy still determines when the agent may act. An autopilot preference, for example, does not itself grant permission to modify protected files.

\paragraph{Expertise and background.}
Profiles needing code guidance account for 35\% of Full and 34\% of Lite; business/design backgrounds account for 25\% and 26\%. Explicitly stated low or high expertise, domain/tool familiarity, advanced review, and source-evidence literacy remain separate categories. They describe different kinds of knowledge rather than positions on a single proficiency scale. This decomposition helps distinguish a need for accessible explanations from an ability to assess a specialized technical argument.

\paragraph{Trust and skepticism.}
Trust is unstated for approximately 76\% of assignments in both collections. Explicit high trust changes from 15\% to 16\%, low trust from 8\% to 7\%, and explicit skepticism from 2\% to 1\%. Skepticism is retained separately from low trust because the recorded descriptors do not establish their equivalence. In particular, the large unstated group cannot be interpreted as neutral or trusting, which limits conclusions about trust-conditioned agent behavior.

\paragraph{Explanation.}
Learning-oriented explanations (35\% $\rightarrow$ 34\%) and operational evidence (25\% $\rightarrow$ 25\%) are the most common preferences. Other profiles reject basic tutorials, request explanations more generally, prioritize outcomes, or seek decision and uncertainty summaries. These categories distinguish what makes an explanation useful: a causal walkthrough may serve a learner, while a concise account of validation and remaining uncertainty may better support a review decision. Explanation quality, therefore needs to be assessed against the disclosed obligation rather than response length alone.

\paragraph{Verification.}
Validation is the largest verification group (34\% $\rightarrow$ 31\%), alongside outcomes/operability, explanation/self-check, specialist/recovery evidence, skipped tests, cost-aware checking, and review evidence. These preferences concern different evidence needs. An explanatory self-check cannot establish an execution result, and a skip-tests preference does not remove the benchmark's native acceptance condition. The task binding must resolve each preference into compatible, explicit obligations and admissible evidence.

\section{Distribution of evolved requirements}
\label{app:requirement-distribution}
\begin{figure}[!htbp]
\centering
\includegraphics[width=\textwidth]{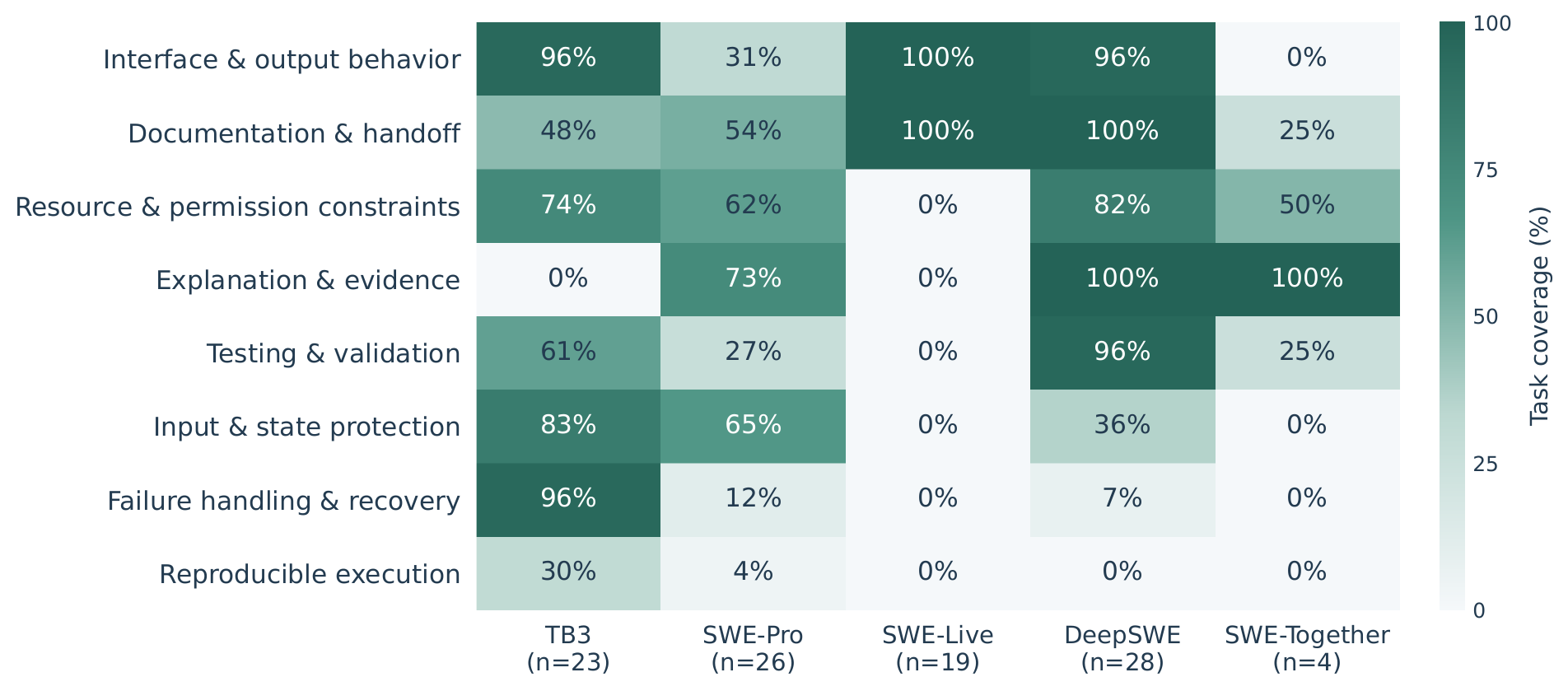}
\caption{Evolved-requirement category coverage in Lite. Each cell is the percentage of tasks in a source subset with at least one added requirement in that category; column labels give denominators. Categories overlap, and displayed percentages are rounded.}
\label{fig:requirement-distribution}
\end{figure}

Evolved requirements specify what a particular user expects beyond the native task contract. Figure~\ref{fig:requirement-distribution} summarizes their content in the Lite subset. The five sources contribute different kinds of obligations, so the pooled distribution should be interpreted together with the within-source coverage.

\subsection{Content Categories}
\label{app:requirement-counting}

Lite contains 186 configured requirement entries across 100 tasks, averaging 1.86 entries per task: 42 tasks have one entry, 30 have two, and 28 have three. An entry is a configured disclosure/scoring unit and may combine multiple atomic obligations. We count the released entries, including event-scheduled entries, rather than observed disclosures, user turns, or Satisfied requirements. Consequently, these counts describe benchmark construction, not interaction length or empirical difficulty.

Each entry receives one or more content labels from an explicit codebook. \emph{Interface and output behavior} covers added API, boundary, schema, or response-format contracts; \emph{documentation and handoff} covers usage documents, runbooks, review inventories, and user-facing delivery. \emph{Resource and permission constraints} covers dependency, resource, network, scope, or approval restrictions. \emph{Explanation and evidence} requires substantive causal explanation, source attribution, or a distinction between observation and inference; a heading such as ``Error'' alone does not qualify. \emph{Testing and validation} includes added tests, probes, comparisons, or explicit validation-status obligations. \emph{Input and state protection} covers user data, credentials, backups, and pre-existing workspace or output state. \emph{Failure handling and recovery} covers failure-safe behavior, restoration, and rollback. \emph{Reproducible execution} requires a reusable or repeatable execution/delivery workflow; a test invocation alone is insufficient.

Coverage is computed at the task level: a task contributes at most once to each category, irrespective of how many of its entries receive that label. Across Lite, interface/output behavior and documentation/handoff are most frequent (76\% and 73\%), followed by resource/permission constraints (58\%), explanation/evidence (51\%), testing/validation (49\%), input/state protection (46\%), failure handling/recovery (27\%), and reproducible execution (8\%). These percentages need not sum to 100\%.

\section{Evaluation Reliability}
\label{app:checker-judge}

A central validity question is whether the coding agent receives the task requirements faithfully and at the intended time. 
Our answer rests on a frozen specification, controller-governed disclosure, and records of agent-visible delivery. 
The simulator supplies conversational language around that contract. The online requirement evaluator assesses the agent's response to active obligations; the personalization verifier assesses final personalization acceptance. Simulator response validation separately checks what the simulator proposes to send, while the controller applies disclosure and progression rules. We distinguish three properties: \emph{specification coverage} (the retained task information is represented in the frozen contract), \emph{delivery fidelity} (the due information reaches the agent without being changed), and \emph{fulfillment} (the agent satisfies the disclosed obligation). A positive fulfillment judgment alone cannot establish the first two properties.

\subsection{Faithful and Complete Requirement Delivery}
\label{app:requirement-delivery}

\paragraph{Freeze the content before generating the dialogue.}
Each task--persona binding stores the native task information, evolved requirements, disclosure conditions, and acceptance rules before an evaluation episode starts. Construction review checks that the user-facing request and its acceptance rubric express the same obligation, including permissions, timing, exceptions, and evidence requirements. Initial-intent blurring changes when native details are visible; it does not authorize their deletion from the complete contract. Native specifications and evolved requirements remain separately identifiable. This provides a reference against which a delivered message can be checked, rather than relying on the simulator to recall every requirement from a persona description.

\paragraph{Preserve mandatory content during rendering.}
The reviewed profiles use concrete mechanisms to protect this reference content (Table~\ref{tab:delivery-safeguards}). Terminal-bench 3 appends frozen native facts and the due requirement disclosure to the simulator's conversational response; at designated work or completion checkpoints, the host supplies all remaining native facts. SWE-Together requires the due requirement's registered statement to occur character-for-character in the proposed message and rejects a missing or changed disclosure. DeepSWE validates the selected public-document identifiers and expands them into their exact stored bodies; simulator response validation separately reviews the surrounding prose for contradictions, unauthorized work, and premature disclosure. SWE-bench Pro additionally records which native and evolved requirements were delivered and audits their presence in actual agent inputs. These mechanisms constrain the communication channel beyond a prompt asking the model to be complete. They have different coverage across profiles and runtime revisions.

\begin{table}[!htbp]
\centering
\small
\caption{Requirement-delivery safeguards in the reviewed source profiles. These are implementation mechanisms, not a measured benchmark-wide semantic fidelity rate.}
\label{tab:delivery-safeguards}
\renewcommand{\arraystretch}{1.15}
\begin{tabularx}{\linewidth}{@{}P{2.5cm}Y@{}}
\toprule
\textbf{Profile} & \textbf{Protection of the delivered contract} \\
\midrule
Terminal-bench 3 & Host assembly appends frozen native facts and the due persona disclosure; completion is gated on the required disclosure and progress state. \\
SWE-bench Pro & Native wording inventory and disclosure ledger; delivery audit checks recorded agent inputs or mini-agent trajectory receipts, including missing or late disclosures. \\
SWE-bench-Live & Frozen opening followed by the complete original issue on user turn 2; the persona handoff contract is fixed separately. \\
SWE-Together & Exact inclusion of the due persona statement; invalid proposals are retried, and acceptance requires all persona requests to be disclosed and assessed as Satisfied. Native follow-ups remain context-dependent. \\
DeepSWE & Frozen staged native units and persona requests; validated document selection and host expansion preserve required public blocks, with simulator response validation of surrounding user prose. \\
\bottomrule
\end{tabularx}
\end{table}

\paragraph{Check delivery at the agent boundary.}
A requirement in a simulator prompt or an unsent draft has not yet been communicated to the coding agent. Delivery evidence must identify the agent-visible message, its content, and its position in the interaction. For example, the SWE-bench Pro audit matches visible user messages against recorded agent calls or mini-agent trajectory receipts, checks source wording and message hashes, and flags missing disclosures or disclosure without the prescribed response opportunity. The controller also distinguishes disclosure from implementation authorization: mentioning a later approval step does not grant that approval. Only after a permitted message is delivered can it update the visible requirement or authorization state.

Completeness is relative to the frozen disclosure policy. At each checkpoint, all information due under that policy must have been supplied before work that depends on it, with the prescribed opportunity to respond. A successful completion must not bypass a mandatory disclosure. An episode may nevertheless stop on an Unresolved current obligation or exhaust its valid budget before a later request becomes eligible; that is different from a simulator omitting an already due requirement. Missing required delivery or illegal progression is an evaluation-validity issue, while failure to satisfy a validly delivered obligation concerns agent performance. The source-specific audits and stop reasons must be retained to distinguish these cases. We do not assume a single identical audit implementation across all six profiles.

\paragraph{What the available evidence establishes.}
The information-disclosure audit provides two direct checks. For DeepSWE, the stored native units reconstruct the original description up to whitespace for all 64 Full tasks; this checks the frozen decomposition, not whether every stage was reached in a particular run. For SWE-bench-Live, all 45 original issues are nonempty and match their configured runtime assets. In 60 historical completed task--persona trials covering those 45 originals, the complete original issue occurs verbatim in user turn 2 in every inspected transcript. These observations support preservation and delivery of the audited native text; they exclude runs interrupted before that turn and do not establish complete semantic communication across every release episode. Appendix~\ref{app:intent-disclosure} details the opening and staging audit, and Appendix~\ref{app:eval-flow} describes the delivery and validity boundaries.

\paragraph{Scope of the guarantee.}
Exact insertion or matching protects the specified text from omission or rewriting within its checked scope. It does not prove that the agent understood the request, that every free-form clarification faithfully expresses the original intent, or that an LLM judge used for simulator response validation always detects contradictory surrounding prose. In particular, SWE-Together's conditional native follow-ups and semantic feedback still involve model judgment. A benchmark-wide semantic completeness claim would require transcript-level annotation of expected versus delivered obligations, including omissions, altered constraints, and added demands. The present evidence supports an auditable delivery protocol with bounded checks; the following analyses test deterministic checker behavior and LLM-judge consistency in the personalization verifier. They do not measure semantic simulator response validation accuracy.

\begin{figure}[!htbp]
\centering
\includegraphics[width=\textwidth]{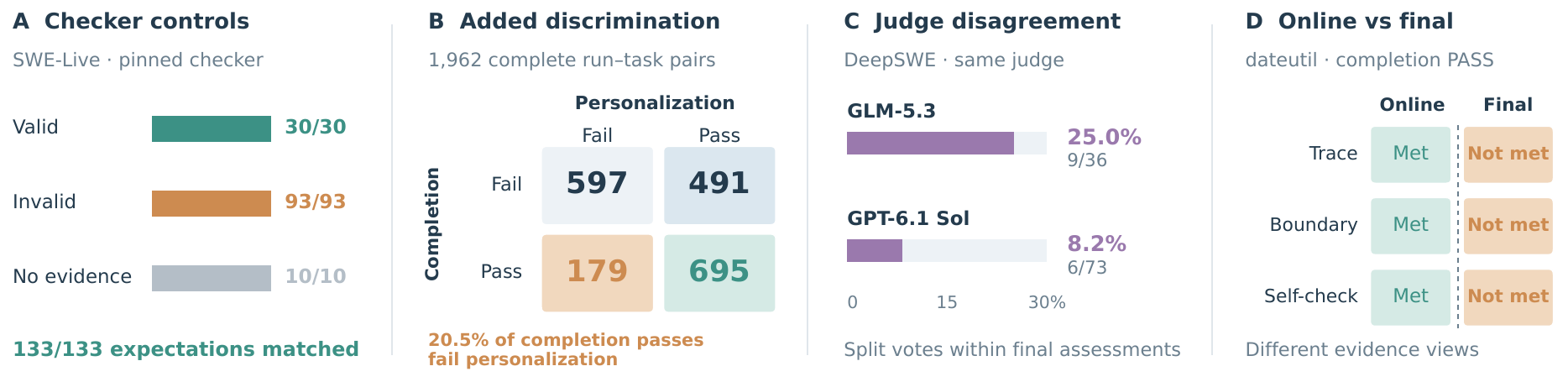}
\caption{Checker effectiveness and judge consistency. (A) The pinned SWE-bench-Live handoff checker matches all 133 specified control outcomes; missing evidence correctly yields \textit{Unresolved}. (B) Among 1,962 complete run--task records, 179 of 874 native-test passes fail persona assessment. (C) Final three-vote DeepSWE assessments disagree in 9/36 and 6/73 decisions for two coding-agent trajectory sets, with the same grading model. (D) A dateutil trajectory receives three online \textit{Satisfied} judgments at turn 8 but three final \textit{Unmet} judgments. The online requirement evaluator and personalization verifier observe different evidence. Each panel has its own denominator and scope, detailed below.}
\label{fig:checker-judge-reliability}
\end{figure}

Figure~\ref{fig:checker-judge-reliability} summarizes four further audits: deterministic contract enforcement, outcome discrimination, within-assessment voting consistency, and online versus final judgments. Each has its own population and does not establish benchmark-wide, human-validated judge accuracy.

\subsection{Deterministic Checker Effectiveness}
\label{app:checker-controls}

Panel A replays targeted construction-audit fixtures against the pinned SWE-bench-Live literal handoff-format checker, rather than reusing historical checker verdicts. Across ten persona templates, all 133 outcomes match their specifications: 30 valid handoffs are accepted, 93 invalid handoffs are rejected, and ten missing-evidence inputs return \textit{Unresolved}. Positive controls include truthful failure reports and the 180-word boundary. Negative controls cover the 181-word boundary, missing or extra bullets, incorrect order or labels, empty content, whitespace violations, and three manager no-code violations. This tests both acceptance and rejection behavior, including abstention when evidence is absent.

As a separate compatibility check, all 37 available SWE-bench-Live episode contracts in the two inspected standard runs match the pinned disclosure and checker contract. Replaying the 35 episodes with paired final checker reports reproduces every verdict and component flag; the other two lack paired reports. These results establish enforcement of this bounded format contract. They do not validate the truth of a handoff's claims, arbitrary SQL semantics, native repair correctness, or all checkers across the benchmark. The 133 targeted fixtures are not independent sampled tasks.

\subsection{Additional Outcome Discrimination}
\label{app:checker-discrimination}

Panel B considers $26\times99=2{,}574$ selected run--task records. We retain 1,962 non-provisional pass/fail records with both completion and personalization grades present and verification not explicitly incomplete. The remaining 612 are excluded in mutually exclusive priority order: 316 invalid or infrastructure records, 175 incomplete verifications, and 121 records missing a component grade. We retain 213 interaction-timeout records with complete component grades, following the recorded scorer's completeness semantics.

The Joint pass outcomes are 597 completion fail/personalization fail, 491 completion fail/personalization pass, 179 completion pass/personalization fail, and 695 completion pass/personalization pass. Thus, $179/874=20.5\%$ of complete native-test passes fail the recorded persona criteria. Persona assessment therefore distinguishes outcomes that native correctness alone would merge. This is descriptive discrimination, not evidence that all 179 judgments are correct or that evolved requirements causally reduced performance. The recorded personalization acceptance outcome can combine deterministic checks, LLM judgments, and online requirement evaluation. Repeated tasks across models and reasoning settings are dependent observations, and exclusions, retry selection, and runtime differences can affect the distribution.

\subsection{Within-Assessment LLM Judge Consistency}
\label{app:judge-consistency}

Panel C uses final DeepSWE checks from two high-effort coding-agent runs, GLM-5.3 and GPT-6.1 Sol, with 28 scored task attempts per run. Both use GPT-6 Luna at medium effort as the LLM judge within the personalization verifier. The unit is a recorded final requirement decision with three resolved votes on its frozen evidence record. Protocol-repair attempts do not count as extra votes. A split occurs when the three recorded statuses are not all equal.

For GLM-5.3 trajectories, 36 of 53 recorded requirement decisions have three votes: eight are unanimously \textit{Satisfied}, 15 unanimously \textit{Unmet}, four unanimously \textit{Unresolved}, and nine split. For GPT-6.1 Sol trajectories, 73 of 76 decisions have three votes: 61, four, and two are unanimous in those respective categories, and six split. The split rates are therefore $9/36=25.0\%$ and $6/73=8.2\%$. The other 20 decisions lack three-vote assessments and are excluded, rather than counted as unanimous; unreached requirements are not imputed.

These rates measure recorded voting consistency, not agreement with human ground truth. Unanimous \textit{Unresolved} is not evidence of semantic correctness. The coding agents produce different evidence and reach different requirements, so the comparison does not establish that the judge is intrinsically more reliable for one agent. Separate complete API-input hashes are not retained for each vote; byte-identical prompts cannot be certified. Documented calibration resampling, up to eight attempts, and selected retries also make this a conditional audit of retained assessments.

\subsection{Online Versus Final Evidence: A Trajectory Case}
\label{app:judge-trajectory}

Panel D follows the GLM-5.3/high attempt \nolinkurl{lite-cpu-20261016-main-glm-5.3/0076} for \nolinkurl{deepswe/dateutil-rfc5545-timezone-interop--student_needing_explanation}. It terminates by final delivery and passes native verification. At online turn 8, the online requirement evaluator marks the counterexample trace, patch-and-boundary reasoning, and self-check requirements \textit{Satisfied}. The personalization verifier marks all three \textit{Unmet}, with respectively one, zero, and one \textit{Satisfied} vote out of three, producing a failed persona outcome.

LLM-judge rationales from the personalization verifier mention unsupported execution claims, incorrect source explanations, and stale references to earlier patch or validation records. These are model-generated explanations, not independently adjudicated defect labels. The online requirement evaluator and personalization verifier consume different evidence views, so the discrepancy alone establishes neither an online error nor a correct final verdict. The case illustrates why native tests, conversational progress, and final evidence-grounded acceptance remain separate. Establishing semantic judge accuracy would additionally require independently annotated cases, human agreement measurements, and adjudication of disputed verdicts; neither controlled format checks nor voting consensus substitutes for that validation.

\subsection{Reliability of Online Requirement Evaluation}

To evaluate the reliability of our requirement evaluation mechanism, we conduct a blinded expert review of 148 requirement-level judgments sampled from 15,100 recoverable assessment records across six execution profiles, including 126 online and 22 final assessments. Two software engineering experts independently annotate each case as \textit{Satisfied}, \textit{Unmet}, or \textit{Unresolved}, with disagreements resolved through a separate adjudication process. After correcting for stratified sampling, the assessments achieve 71.0\% agreement with the adjudicated expert reference (Cohen's $\kappa=0.473$). When reduced to the operational accept/reject decision, agreement increases to 77.2\%, with a Matthews correlation coefficient of 0.541. Notably, \textit{Satisfied} judgments achieve 89.2\% precision and 74.5\% recall, indicating that most accepted requirements are also supported by expert review. Although these results demonstrate meaningful alignment between automated assessment and human judgment, the remaining disagreements highlight the importance of explicit evidence boundaries and uncertainty handling. Overall, the study provides empirical support for the evaluation mechanism as a scalable, albeit imperfect, approach to evaluating user-specific requirement fulfillment.

\section{Initial-Intent Blurring and Progressive Requirement Disclosure}
\label{app:intent-disclosure}

Our adaptation separates ``what the software must do'' from ``how this user expects the work to be performed and delivered''. The native task defines the original software objective. Initial-intent blurring changes how much of that task is stated in the opening request, while evolved requirements introduce explicit, persona-grounded obligations concerning the workflow, evidence, or handoff. Revealing an omitted native API contract is therefore not itself an evolved requirement. Both the original task and the evolved requirements are frozen before evaluation; progression changes their visibility and activation, rather than inventing new acceptance criteria during a run.

\subsection{Constructing the Opening Request}
\label{app:intent-opening}

The construction records support four opening mechanisms (Table~\ref{tab:intent-mechanisms}). For summarized tasks, the opening expresses the task-specific goal while omitting selected implementation or acceptance details, such as input paths, output formats, numerical limits, and boundary behavior. Deferred details remain grounded in the native specification. For staged tasks, the native instruction is partitioned into units with predetermined disclosure stages. These mechanisms are distinct from the complete issue deferral used by SWE-bench-Live and from profiles that retain the native description upfront.

\begin{table}[!htbp]
\centering
\small
\caption{Opening mechanisms in the 231-task Full collection. Counts refer to task--persona pairs, with one selected persona per original task.}
\label{tab:intent-mechanisms}
\renewcommand{\arraystretch}{1.15}
\begin{tabularx}{\linewidth}{@{}P{3.0cm}rY@{}}
\toprule
\textbf{Mechanism} & \textbf{Tasks} & \textbf{Recorded construction} \\
\midrule
Concise summary & 75 & Terminal-bench 3: 52; SWE-bench Pro: 23. The opening summarizes the native goal and defers details. \\
Complete issue deferral & 45 & SWE-bench-Live. A generic repository-help request precedes the complete original issue on user turn 2. \\
Staged native disclosure & 64 & DeepSWE. S1 native units appear initially; later stages disclose the remaining native units. \\
Native description upfront & 47 & SWE-bench Pro: 38; SWE-Together: 9. Later interaction can still introduce evolved requirements or native follow-ups. \\
\bottomrule
\end{tabularx}
\end{table}

The SWE-bench-Live opening asks for help with a repository problem without specifying the issue. All 45 original issues remain present in the frozen task assets and are configured for the second user turn. A separate audit of 60 historical completed task--persona trials, covering these 45 originals, found the original issue verbatim in user turn 2 in every audited trial. This is evidence about those completed historical runs, not a fresh execution of every release task or a guarantee for runs interrupted before turn 2.

\subsection{Measuring Task-Level Blurring}
\label{app:intent-measurement}

Figure~\ref{fig:task-blurring} reports each task's native-description shortening. Let $w_i^{\mathrm{native}}$ count whitespace-delimited words in the original native description, and let $w_i^{\mathrm{opening}}$ count its task-specific opening representation. We plot
\begin{equation}
 b_i = \left(1-\frac{w_i^{\mathrm{opening}}}{w_i^{\mathrm{native}}}\right).
\end{equation}
Thus, reducing a 500-word description to a 50-word opening yields $b_i=90\%$. This is a text-length measure, not the percentage of semantic information concealed or of the core goal lost. In particular, SWE-bench-Live has $b_i=100\%$ because its generic opening contributes zero task-specific native words; the complete issue is subsequently disclosed. An upfront task can have $b_i=0\%$ and still contain substantial persona-driven interaction.

The text pair is source-specific. Terminal-bench 3 uses the frozen task-only construction request before persona styling and removes benchmark canary comments from the original. DeepSWE uses S1 native units; all native units together reconstruct the original up to whitespace. SWE-bench Pro counts its issue statement and requirements, excluding the separate interface field and known explicit persona-only preservation sentences. SWE-Together uses its native instruction verified in the opening. These definitions exclude persona and runtime scaffolding from the measured native text. Detail labels illustrate deferred items rather than exhaustive categories or importance weights.

\subsection{Introducing evolved requirements During Interaction}
\label{app:intent-progression}

Each frozen evolved requirement specifies the obligation and its assessment contract; the profile additionally determines when it is eligible for disclosure and what evidence or event permits progression. The controller exposes the current request to the agent, uses the online requirement evaluator's verdict on eligible agent evidence, and either continues the active work, requests a repair or clarification, or delivers the next permitted request after simulator response validation. The simulator expresses authorized feedback in the persona's style. It cannot make an undisclosed future obligation active merely by proposing it, and it cannot reveal hidden tests or reference solutions. Native task facts, evolved requirements, and permission to edit are maintained separately.

The progression rule is not identical across sources. Terminal-bench 3 and Pro-mini use requirement-gated interaction around a frozen opening; Pro-shell uses event-based persona disclosure with the native contract initially available. SWE-bench-Live schedules its full issue at turn 2 and assesses its frozen persona handoff contract. SWE-Together preserves native follow-ups alongside additional persona requests. DeepSWE follows a frozen S1--S4 disclosure and authorization plan, whose stages are logical states rather than four fixed conversation turns. A run may spend multiple turns repairing one stage or stop before later requests are delivered. Section~\ref{app:eval-flow} specifies the profile-dependent execution and evaluation order.

For a concrete example, the DeepSWE \path{abs-module-cache-flags} task pairs ABS module-loading work with a first-time PR contributor. Its initial native units describe module search paths and canonicalization. The initial interaction requests a short plan and reserves implementation authorization. The configured progression then combines the remaining native specifications with three distinct evolved requirements:
\begin{enumerate}[leftmargin=*,itemsep=3pt]
  \item \textbf{S1: scoped work and preservation.} Keep the patch within the allowed source paths, preserve user-owned workspace content, and provide a scope manifest at completion.
  \item \textbf{S2: explanation grounded in evidence.} Disclose the native cache visibility and reset contracts, and request an explanation of the public module-path case based on baseline observations and source evidence.
  \item \textbf{S3: reviewable delivery.} Disclose the remaining native cycle, tracing, and CLI behavior, and request a PR draft and validation record tied to the delivered revision.
  \item \textbf{S4: final review.} Check the accumulated obligations without adding another evolved requirement.
\end{enumerate}
This example describes the configured plan. The historical trajectory in Section~\ref{app:case-study} illustrates why the plan must be distinguished from what was actually reached: unresolved evidence can block a later handoff request. Evaluation therefore records delivered requests, active obligations, authorization, and stop reasons, rather than treating every configured stage as completed interaction.

\begin{figure}[p]
\centering
\includegraphics[width=\textwidth,height=.88\textheight,keepaspectratio]{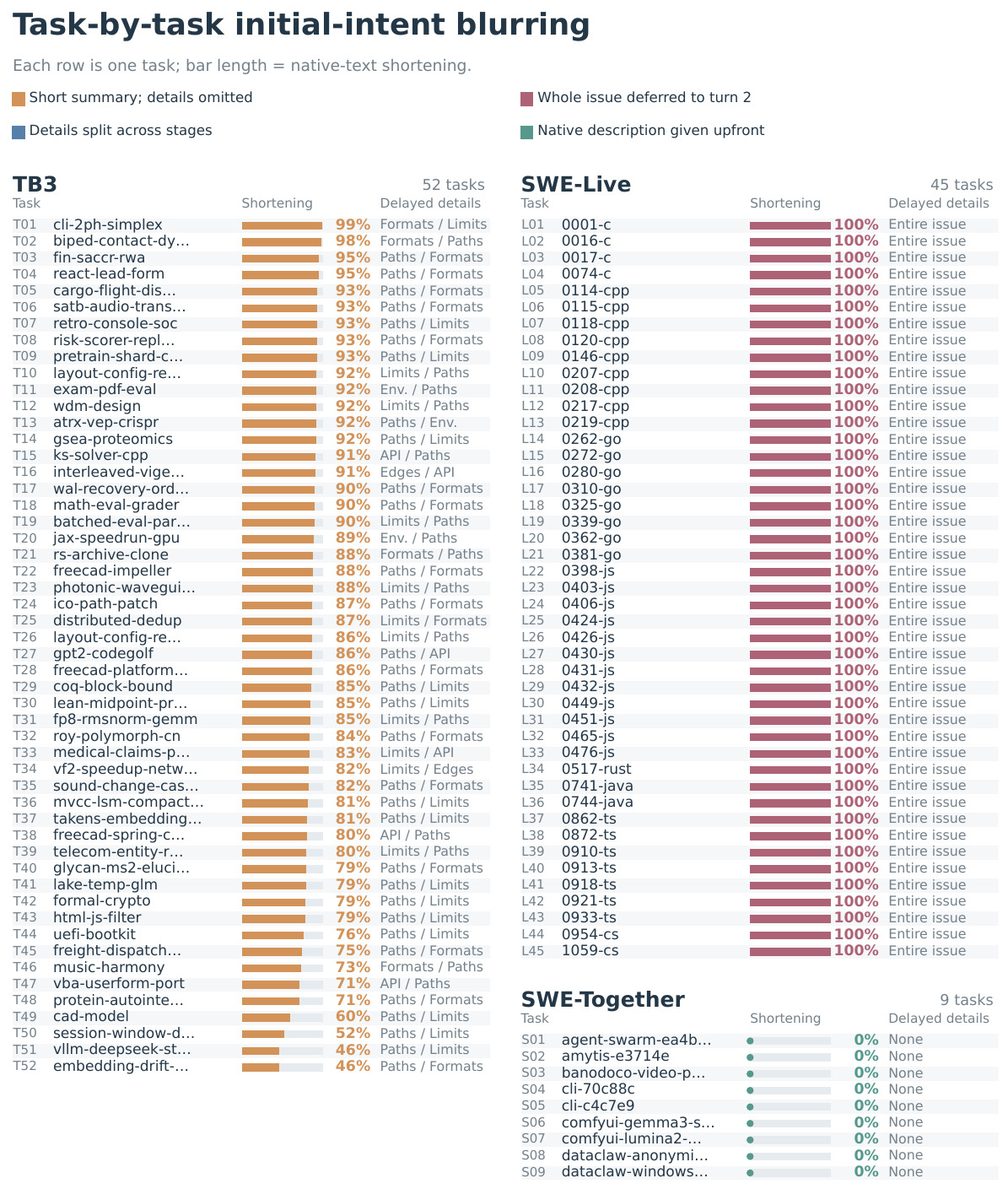}
\caption{Task-level initial-intent blurring, shown over two pages for readability. Part 1 contains Terminal-bench 3, SWE-bench-Live, and SWE-Together. Bars show native-description shortening $b_i$; colors identify the opening mechanism. Right-hand labels give examples of delayed native details. The measure concerns text length, not semantic information loss.}
\label{fig:task-blurring}
\end{figure}
\clearpage
\begin{figure}[p]
\ContinuedFloat
\centering
\includegraphics[width=\textwidth,height=.88\textheight,keepaspectratio]{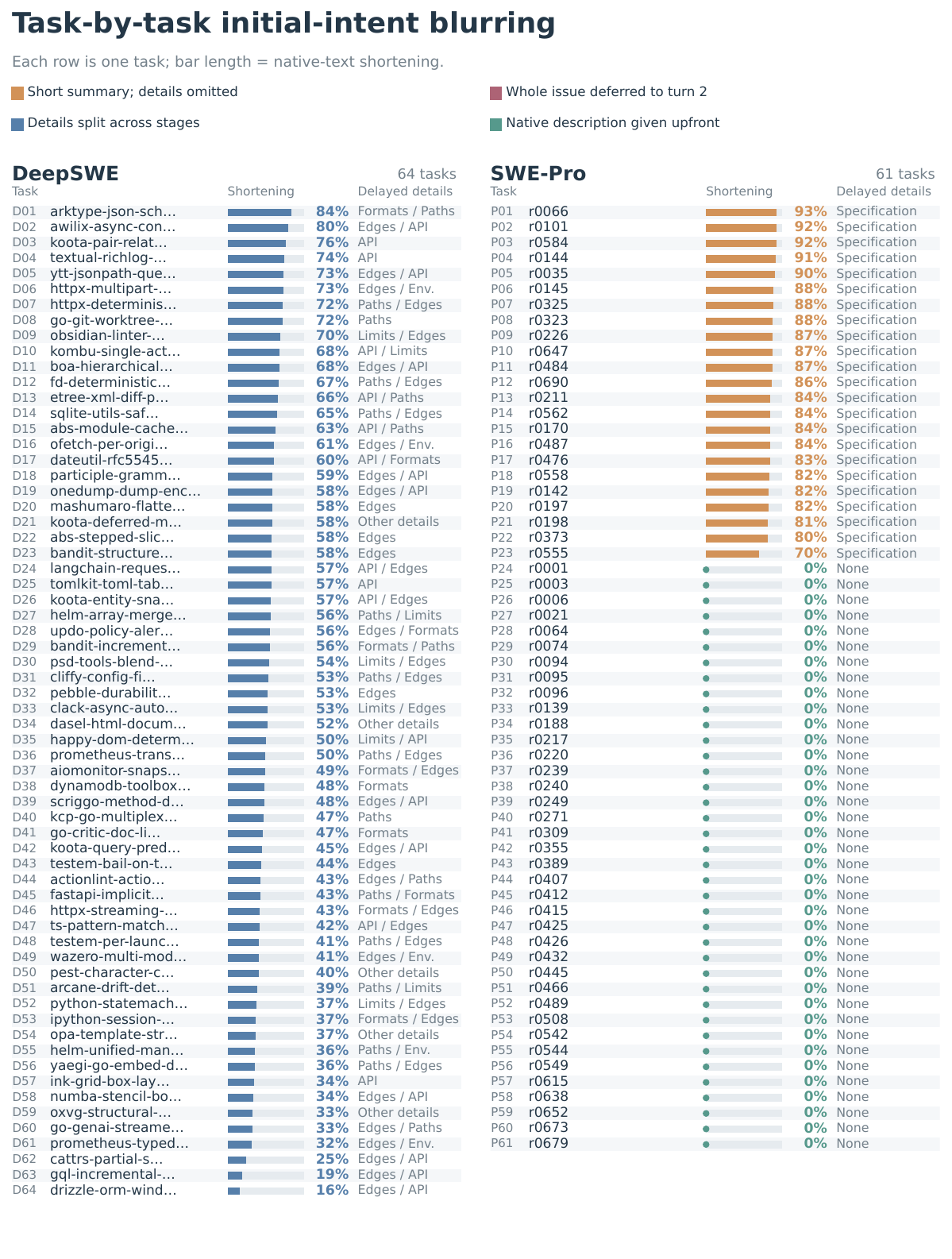}
\caption{Task-level initial-intent blurring (continued): DeepSWE and SWE-bench Pro. All 231 Full tasks are shown across the two pages. Zero shortening is scoped to the native text definition and does not imply the absence of subsequent evolved requirements.}
\end{figure}

\section{Additional Experimental Results}
\label{app:additional-results}

\subsection{Performance on Common Non-Timeout Tasks}
\label{app:non-timeout}

Valid timeouts receive no joint credit under the budget-aware evaluation. A model can therefore obtain a low overall score despite performing well on the tasks it finishes. We examine this distinction through conditional pass rates on common non-timeout tasks, alongside their sensitivity to effort and time budget.

\paragraph{Common-task evaluation.}
For each model and comparison ladder $\mathcal{L}$, we retain the common tasks that every setting finishes within its time limit with a graded result. If $\mathcal{V}_{c}$ is that task set for setting $c$, we compute
\begin{equation}
 \mathcal{C}_{\mathcal{L}}=\bigcap_{c\in\mathcal{L}}\mathcal{V}_{c},\qquad
 R(c\mid\mathcal{C}_{\mathcal{L}})=
 \frac{\sum_{i\in\mathcal{C}_{\mathcal{L}}}\mathbf{1}[c\text{ passes }i]}{|\mathcal{C}_{\mathcal{L}}|}.
 \label{eq:non-timeout-common}
\end{equation}
The denominator is fixed within each source ladder. The figure combines the full effort ladder at $1\times$ with the max-budget ladder at $2\times$ and $3\times$, which may use a different common-task set. Panel labels report $n$ for each budget group: this is an intersection size, not a per-setting completion count.

\begin{figure}[H]
\centering
\includegraphics[width=\textwidth]{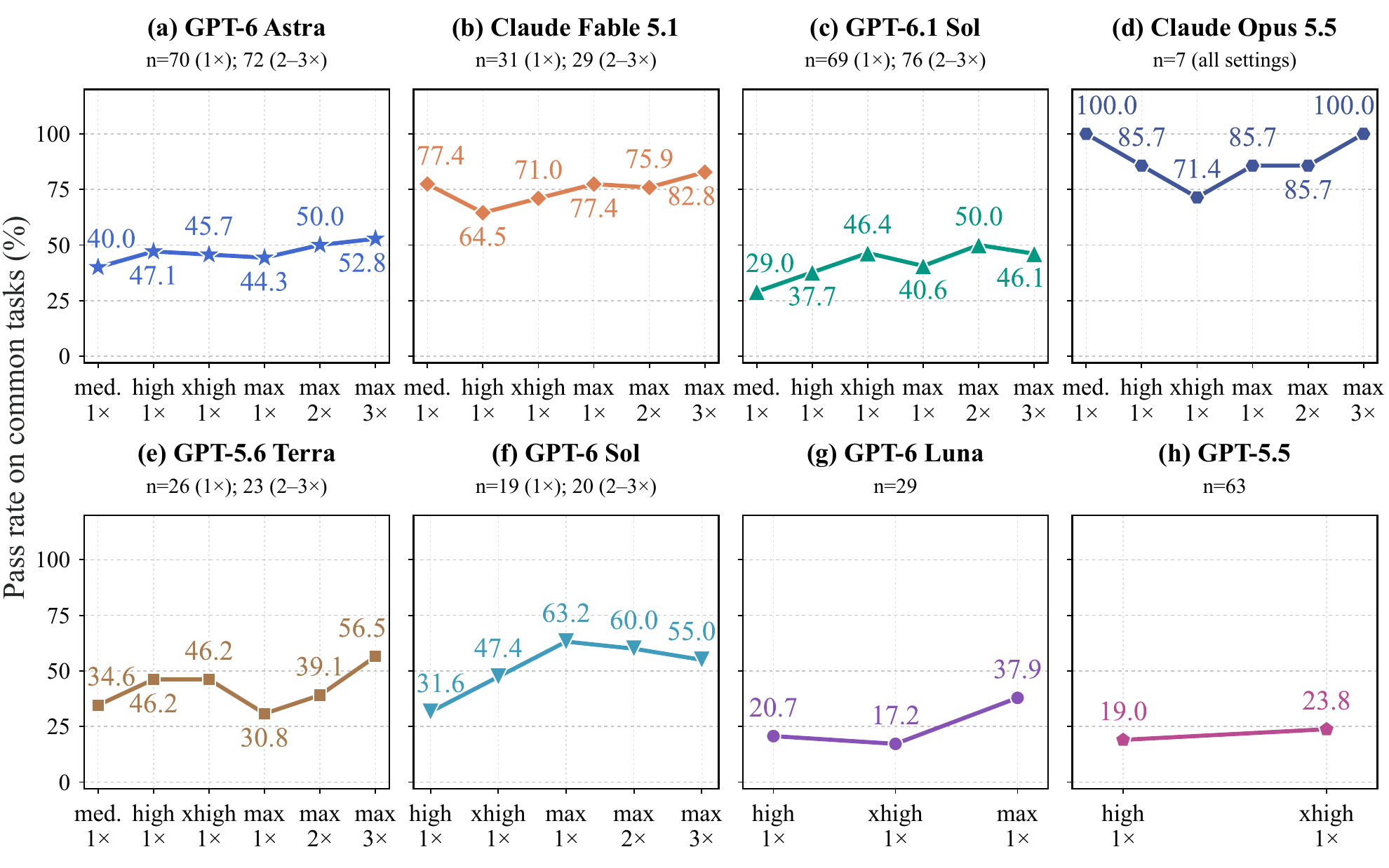}
\caption{One continuous line per model connects all available settings, including max at $1\times$, $2\times$, and $3\times$. Every setting has one point labeled with its pass rate (\%). Rates at $1\times$ come from the full effort ladder, including its max-$1\times$ value; max-$2\times$ and max-$3\times$ rates come from the time-budget ladder. Lines are visual guides: the task set can change between $1\times$ and $2$--$3\times$, so a connected series is not necessarily a matched-task comparison throughout. Claude Opus 5.5 uses the same seven tasks across all six settings. Med. denotes medium effort.}
\label{fig:non-timeout-scaling}
\end{figure}

\paragraph{Conditional performance and selection.}
At max effort and $1\times$ budget, Claude Fable 5.1 passes $77.4\%$ of its 31 common tasks, and Claude Opus 5.5 passes $85.7\%$ of its seven common tasks. These high conditional rates illustrate that models can succeed on a large fraction of retained tasks even when timeouts reduce their overall scores. However, we cannot rule out that the tasks they finish are easier or better matched to their capabilities than those that time out. Completion is a selection criterion, so these rates do not estimate success on the excluded tasks or establish stronger performance across the full benchmark. The model-specific task sets also prevent a controlled cross-model ranking.

\paragraph{Effort scaling.}
Six of eight models exhibit at least one decrease as effort increases at the fixed $1\times$ budget on their respective common-task sets. GPT-6.1 Sol rises from $29.0\%$ at medium to $46.4\%$ at xhigh, then falls to $40.6\%$ at max on the same 69 tasks. Conversely, GPT-6 Sol increases from $31.6\%$ at high to $63.2\%$ at max on 19 tasks. Thus, higher effort does not consistently yield higher conditional pass rates in these observations.

\paragraph{Time-budget scaling.}
At max effort, extending the budget from $2\times$ to $3\times$ increases the observed rate for four of the six models with both settings and decreases it for two. GPT-6 Astra rises from $50.0\%$ to $52.8\%$ ($n=72$), while GPT-6.1 Sol falls from $50.0\%$ to $46.1\%$ ($n=76$) and GPT-6 Sol falls from $60.0\%$ to $55.0\%$ ($n=20$). Each comparison holds its task set fixed. In contrast, the plotted $1\times$-to-$2\times$ segment can cross task sets and does not isolate the effect of additional time.

\paragraph{Statistical uncertainty.}
A complementary analysis of 33 within-model setting pairs uses two-sided paired sign tests on discordant task outcomes, with a separate common-task set for each pair. Only GPT-5.6 Terra's high-to-xhigh comparison reaches nominal $p<0.05$ ($p=0.049$, $n=81$); none meets the Bonferroni threshold of $0.05/33\approx0.0015$. These pairwise task sets can differ from the plotted ladder sets, so their tests do not establish significance for the displayed changes. Failure to reject also does not establish equivalence between settings.

The seven-task Claude Opus 5.5 subset is particularly sensitive to individual outcomes: one task changes the rate by about 14.3 percentage points. Its $100\%$ values therefore provide limited evidence about general performance. Taken together, these results describe success conditional on completion and grading; they neither remove the practical cost of timeouts nor show that higher effort causes lower performance. The budget-aware Joint pass rate remains the primary full-set performance measure.

\section{Case Study}
\label{app:case-study}
\subsection{DeepSWE: A First-Time Contributor and a Blocked Handoff}
This case follows \path{abs-module-cache-flags} with the \path{first_time_pr_contributor} persona. The coding model is \texttt{GPT-6-luna}; the simulator and semantic judge use \texttt{GPT-6-astra}. We report the historical execution recorded at source revision \texttt{4e7a1469}, separately from the release and runtime identities used elsewhere in this paper. The task is to harden ABS module resolution, caching and script flags. The user asks for a small, explainable patch while protecting pre-existing workspace content.

\begin{tcolorbox}[csbase,colback=csGray!6,colframe=csGray,title={Reading the evidence}]
The Pivot episode contains \textbf{35 agent calls} (24 tools and 11 replies), \textbf{23 simulator calls} (12 renderings and 11 proposals), 32 interaction-time LLM-judge calls covering online requirement evaluation and simulator response validation, and six personalization verifier LLM-judge calls. There are 12 user turns. The source export retains complete visible messages, action payloads, command results, simulator contexts, proposals and controller receipts. It does \emph{not} retain the agent's raw per-call request/response envelopes or provider call IDs. Tool results shown in the case follow the recorded 24,000-character suffix limit. Missing request envelopes and internal reasoning are not reconstructed as recorded model inputs.

\end{tcolorbox}

\paragraph{Disclosure and authority.}
S1 initially permits inspection and a plan only. User turn 2 discloses the isolation obligation and authorizes edits and local commits. S2 arrives at turn 3 after the first patch: it discloses cache visibility/reset and requests a causal explanation of the public \texttt{module-path-v1} example. S3 would add cycle handling, debug tracing, detailed CLI behavior and reviewable PR materials; S4 is the final stage. Neither S3 nor S4 is reached. A feature mentioned in the opening is not equivalent to disclosure of its later detailed acceptance conditions.

\begin{table}[htbp]\centering\small
\caption{Recorded outcomes for the selected pair. Public validation and hidden native acceptance have different scopes.}\label{tab:case-outcome}
\begin{tabularx}{\linewidth}{@{}lYY@{}}\toprule
 & \textbf{Baseline} & \textbf{Pivot}\\\midrule
Task disclosure & Full at start & Progressive; stops at S2\\
Agent calls / tools / replies & 18 / 17 / 1 & 35 / 24 / 11\\
User turns & 1 & 12\\
Hidden fail-to-pass tests & 19/20 & 7/20\\
Hidden pass-to-pass tests & 3/3 & 3/3\\
Native full acceptance & False & False\\
Stop reason & Agent done & User-turn budget exhausted\\
Final persona findings & Not evaluated & Isolation: Satisfied; explanation: Satisfied; handoff unreached\\\bottomrule
\end{tabularx}\end{table}

\paragraph{What the case demonstrates.}
The pivot patch passes the registered public check, but the online requirement evaluator isolation judgment remains Unresolved because the visible initial snapshot lacks the protected file's content identity. The simulator proposes advancement five times after turn 7; the controller rejects every proposal. The personalization verifier later marks isolation and explanation as Satisfied, explicitly disagreeing with the online requirement evaluator isolation judgment. The evidence supports a progression bottleneck and an online/final discrepancy, not a claim that the protected file was actually damaged. The unreached handoff must not be represented as a delivered user request. Likewise, this selected pair does not establish a general model ranking or a causal effect of personalization.

\subsection{User--Agent Conversation}
\label{app:case-timeline}
Left-aligned green bubbles quote the simulator's model-authored user prose; right-aligned blue bubbles reproduce every recorded agent reply. Avatars identify the two speakers; turn markers track the interaction and disclosed stage. Repeated environment boilerplate and host-expanded specification blocks are omitted from the printed dialogue for readability. Indented gray tool notes and amber controller updates are editorial annotations, not additional dialogue. Long messages continue across pages under the same speaker label. IDs A01--A35 identify agent calls; S01--S23 identify simulator calls. The dialogue labels also identify the recorded API call number. In particular, turn 2 ends at a tool checkpoint without a textual agent reply.

\csTurn{1}{S1}

\begin{tcolorbox}[csuser,title={\csUserTitle{1}{S01 / API 1}}]
This is my first PR; please inspect the baseline and the task details disclosed so far, then explain a short implementation plan in plain language. Please leave existing workspace changes untouched and wait for my next request before editing repository source or committing.
\end{tcolorbox}

\begin{tcolorbox}[csinternal,title={\faTerminal\enspace Tool activity | A01, A02, A03, A04}]
A01 / API 3: Read-only status, file inventory and module symbols; exit 0.

A02 / API 4: Read environment, require/cache, runner and REPL source; exit 0.

A03 / API 5: Inspect path helpers, existing tests and Git history; exit 0.
\end{tcolorbox}

\begin{tcolorbox}[csagent,title={\csAgentTitle{A04 / API 6}}]
Baseline inspection complete. I left the repository source, branch, and commit unchanged; the only existing workspace item remains the untracked \texttt{persona-user-notes.txt}, untouched.

Short implementation plan for the disclosed requirements:\newline{} 1. Add a single path-normalization and lookup flow for \texttt{require()}: recognize bare names as \texttt{<name>/index.abs}, search the executing file's directory first, then normalized \texttt{ABS\_MODULE\_PATH} directories in order, and canonicalize the selected file before caching it.\newline{} 2. Parse quoted \texttt{ABS\_MODULE\_PATH} entries, clean/expand them, and remove equivalent directories without changing their first-seen order.\newline{} 3. Refactor module cache handling to use canonical file identities so alternate spellings reuse one entry, while preserving failed-load behavior.\newline{} 4. Inspect the remaining CLI/script-mode behavior before implementation, then integrate module-related flags without changing unrelated REPL behavior.\newline{} 5. Add focused tests only within the eventual allowed source scope as appropriate, run the available checks, and commit the finished task on a new branch from \texttt{main} when implementation is authorized.

I will not edit repository source, create the task branch, or commit until your next request supplies the remaining specifications/authorization to proceed.
\end{tcolorbox}

\begin{tcolorbox}[csdecision,title={\faSlidersH\enspace controller update}]S02 / API 7: simulator proposes start\_implementation; controller accepts; delivered outcome is start\_implementation.\end{tcolorbox}

\csTurn{2}{S1}

\begin{tcolorbox}[csuser,title={\csUserTitle{2}{S03 / API 8}}]
Please begin implementing the module-loading requirements disclosed so far on a new branch from main; you may now edit and commit task source within the permitted scope. Please address the outstanding requirements in the supplied documents, keeping the work limited to the behavior specified so far.
\end{tcolorbox}

\begin{tcolorbox}[csinternal,title={\faTerminal\enspace Tool activity | A05, A06}]
A05 / API 10: Create task/harden-module-loading from the baseline; exit 0.

A06 / API 11: First edit: canonical lookup and module-path parsing in evaluator/functions.go; exit 0. First-patch checkpoint ends the exchange.
\end{tcolorbox}

\textit{No agent prose reply: the first-patch event triggers the next user turn.}\par

\begin{tcolorbox}[csdecision,title={\faSlidersH\enspace controller update}]S04 / API 12: simulator proposes advance; controller accepts; delivered outcome is advance.\end{tcolorbox}

\csTurn{3}{S2}

\begin{tcolorbox}[csuser,title={\csUserTitle{3}{S05 / API 13}}]
Please move on to the newly supplied cache visibility and reset requirements, and use your existing first patch to prepare the requested evidence and plain-language explanation at /deliverables/explanation.md. Please address the outstanding requirements in the supplied feedback while preserving the earlier module-loading behavior, which remains unverified.
\end{tcolorbox}

\begin{tcolorbox}[csinternal,title={\faTerminal\enspace Tool activity | A07, A08, A09, A10, A11, A12, A13, A14, A15, A16, A17, A18, A19}]
A07 / API 15: Inspect hash and builtin representations; exit 0.

A08 / API 16: Read number/hash types and builtin registration; exit 0.

A09 / API 17: Read remaining builtin registration and imports; exit 0.

A10 / API 18: Add cache introspection/reset and format the patch. Broad Go test command fails (exit 1).

A11 / API 19: Create public-case fixtures and a detached baseline worktree. Baseline build fails: invocation is outside the main Go module (exit 1).

A12 / API 20: Correct baseline build; compare the same public case. Old run: exit 99, missing entry/demo/index.abs. New run: exit 0, stdout 42, empty stderr.

A13 / API 21: Write the public-case explanation under /deliverables; exit 0.

A14 / API 22: Run focused tests and a cache probe. Shell exits 0, but nested go\_test\_rc=1; probe exits 0 with formatting artifacts in stdout.

A15 / API 23: Run git diff --check and CONTEXT=abs go test ./...; exit 1, with host/js and update-test failures.

A16 / API 24: Commit evaluator/functions.go as ea1a519; 161 insertions and one deletion. Protected notes remain untracked.

A17 / API 25: Write scope.json with baseline, changed path and protected-file declaration; exit 0.

A18 / API 26: Registered clean-commit public validation public-001: passed. This is not hidden acceptance.
\end{tcolorbox}

\begin{tcolorbox}[csagent,title={\csAgentTitle{A19 / API 27}}]
Implemented and committed on new branch \texttt{task/harden-module-loading}.

Commit: \texttt{ea1a519} (\texttt{harden module resolution and require cache introspection})

Changes in \texttt{evaluator/functions.go}:\newline{} - Canonical module resolution with base-directory-first lookup and ordered \texttt{ABS\_MODULE\_PATH} search.\newline{} - Bare module names resolve through \texttt{index.abs} via the existing alias/path behavior.\newline{} - Quoted, expanded, canonicalized, and deduplicated module-path entries.\newline{} - Canonical cache keys, preserving equivalent module paths as one cache entry.\newline{} - Added \texttt{require\_cache\_info()}, \texttt{require\_cache\_keys()}, and \texttt{reset\_require\_cache()}.\newline{} - Tracks cache hits, misses, size, and active-load-stack depth (\texttt{inflight}).

Evidence artifacts written outside the repository:\newline{} - \texttt{/deliverables/explanation.md} --- includes the observed baseline failure and current public module-path-v1 success (\texttt{42}).\newline{} - \texttt{/deliverables/scope.json} --- required base commit, changed task paths, and protected file declaration.

Validation passed on the clean committed patch (\texttt{public-001}). The pre-existing \texttt{persona-user-notes.txt} remains untouched and untracked. No out-of-scope repository paths were changed.
\end{tcolorbox}

\begin{tcolorbox}[csdecision,title={\faSlidersH\enspace controller update}]S06 / API 30: simulator proposes continue; controller accepts; delivered outcome is continue. Blocking requirements: first\_pr\_isolated\_patch, first\_pr\_patch\_explanation.\end{tcolorbox}

\csTurn{4}{S2}

\begin{tcolorbox}[csuser,title={\csUserTitle{4}{S07 / API 31}}]
Please continue with the disclosed work and address the outstanding evidence requirements in the supplied feedback. Please reconcile your explanation with the existing execution records, including the earlier baseline-build and local-test failures, and clarify any evidence needed to support your claimed results.
\end{tcolorbox}

\begin{tcolorbox}[csinternal,title={\faTerminal\enspace Tool activity | A20, A21, A22, A23}]
A20 / API 33: Read scope, explanation and commit/path evidence; exit 0.

A21 / API 34: Rewrite scope and explanation to include baseline-build failure, observed outcomes and test limitations; exit 0.

A22 / API 35: Collect final status, commit paths and artifact hashes; exit 0.
\end{tcolorbox}

\begin{tcolorbox}[csagent,title={\csAgentTitle{A23 / API 36}}]
Updated the evidence artifacts to reconcile the execution record in detail.

\texttt{/deliverables/explanation.md} now includes:\newline{} - The exact initial and final snapshot states.\newline{} - The isolated task diff and committed-path evidence.\newline{} - The first failed baseline build attempt (wrong Go invocation outside the module), followed by the successful corrected baseline build.\newline{} - The actual baseline public-case failure and source excerpt explaining why it searched only \texttt{entry/demo/index.abs}.\newline{} - The current public-case result (\texttt{exit 0}, stdout \texttt{42}, empty stderr).\newline{} - The clean-commit validation record \texttt{public-001} and patch hash.\newline{} - Local test limitations: the initial environment-sensitive failures, the passing \texttt{CONTEXT=abs} evaluator run, and unrelated host/js and update-test failures in the broad run.\newline{} - Clear distinctions between observations, implementation-based predictions, and untested limits.

\texttt{/deliverables/scope.json} now contains the required fields plus initial/final snapshot references, the exact isolated diff command, full commit ID, and preserved-path information.

Final repository evidence remains unchanged: branch \texttt{task/harden-module-loading}, HEAD \texttt{ea1a51955aee5e5d\allowbreak{}d84618a876fc2831\allowbreak{}1bd9233e}, committed task path only \texttt{evaluator/functions.go}, and preserved untracked \texttt{persona-user-notes.txt}.
\end{tcolorbox}

\begin{tcolorbox}[csdecision,title={\faSlidersH\enspace controller update}]S08 / API 39: simulator proposes continue; controller accepts; delivered outcome is continue. Blocking requirements: first\_pr\_isolated\_patch, first\_pr\_patch\_explanation.\end{tcolorbox}

\csTurn{5}{S2}

\begin{tcolorbox}[csuser,title={\csUserTitle{5}{S09 / API 40}}]
Please continue with the disclosed work and address the outstanding evidence requirements in the supplied feedback, which remain unverified. Please show the supporting contents of /deliverables/scope.json and /deliverables/explanation.md rather than only summarizing them, and clarify any gaps in the existing evidence.
\end{tcolorbox}

\begin{tcolorbox}[csinternal,title={\faTerminal\enspace Tool activity | A24, A25}]
A24 / API 42: Print complete scope and explanation artifacts plus path evidence; exit 0.
\end{tcolorbox}

\begin{tcolorbox}[csagent,title={\csAgentTitle{A25 / API 43}}]
The supporting contents are now exposed directly in the execution record above:

- \texttt{/deliverables/scope.json} shows the required values exactly: \texttt{base\_commit} is \texttt{cb1b3b671d0ee9fa\allowbreak{}9da9f7b02f869679\allowbreak{}53ffd10a}, \texttt{new\_files} is \texttt{["evaluator/functions.go"]}, and \texttt{protected\_files} is \texttt{["persona-user-notes.txt"]}. It also includes the initial snapshot, final snapshot, full task commit, and reproducible isolated-diff commands.\newline{} - \texttt{/deliverables/explanation.md} shows the requested case source, baseline source excerpt, corrected baseline-build record, observed baseline failure, current success, patch symbols and causal chain, validation record, local-test records, and explicit scope limits.

Current supporting boundary output shows:

\textbf{task/harden-module-loading}\newline{} ?? persona-user-notes.txt\newline{} M evaluator/functions.go

The baseline-to-HEAD diff contains only \texttt{evaluator/functions.go}; the committed task tree also contains only that modified task path. \texttt{persona-user-notes.txt} remains untracked, with current mode \texttt{664} and current SHA-256 \texttt{3cacf0...962c4c}.

One evidence gap remains and is stated plainly: the initial snapshot record provided the protected file's path/state (untracked, not staged or modified), but did not capture its initial content hash or mode. Therefore the available record proves I did not use or overwrite it and that its Git-visible state was preserved, but cannot independently compare its initial bytes/mode against the final hash/mode. No repository operation in this session edited that file. The task diff and public validation evidence are otherwise directly reproducible and exposed in the artifacts.
\end{tcolorbox}

\begin{tcolorbox}[csdecision,title={\faSlidersH\enspace controller update}]S10 / API 46: simulator proposes continue; controller accepts; delivered outcome is continue. Blocking requirements: first\_pr\_isolated\_patch, first\_pr\_patch\_explanation.\end{tcolorbox}

\csTurn{6}{S2}

\begin{tcolorbox}[csuser,title={\csUserTitle{6}{S11 / API 47}}]
Please continue with the disclosed work and address the outstanding evidence requirements in the supplied feedback, which remain unverified. Please reconcile /deliverables/explanation.md with the existing execution records and clarify any remaining evidence gaps without treating Git-visible state alone as proof of preserved file contents or modes.
\end{tcolorbox}

\begin{tcolorbox}[csinternal,title={\faTerminal\enspace Tool activity | A26, A27}]
A26 / API 49: Append the protected-file evidence gap; record final mode 664, size 89 and SHA-256; exit 0.
\end{tcolorbox}

\begin{tcolorbox}[csagent,title={\csAgentTitle{A27 / API 50}}]
Reconciled \texttt{/deliverables/explanation.md} and exposed its relevant contents, together with the complete \texttt{/deliverables/scope.json}, in the execution record above.

The explanation now explicitly states the remaining preservation gap: the initial host snapshot recorded \texttt{persona-user-notes.txt} as untracked and reserved, but did not record its initial bytes, hash, size, or mode. Therefore no honest initial-to-final content/mode comparison can be supplied. The final observation is recorded (\texttt{mode=664}, size \texttt{89}, SHA-256 \texttt{3cacf0...962c4c}), but I do not treat unchanged Git-visible status as proof of equality.

The available evidence does establish the task boundary: initial HEAD and state, final HEAD and state, isolated diff path \texttt{evaluator/functions.go}, and the required manifest fields. The explanation separately exposes the exact public case, baseline source/failure, corrected baseline build, current result, validation record, local-test limitations, causal mechanism, and untested scope limits. No repository source or protected file was changed during this evidence update.
\end{tcolorbox}

\begin{tcolorbox}[csdecision,title={\faSlidersH\enspace controller update}]S12 / API 53: simulator proposes continue; controller accepts; delivered outcome is continue. Blocking requirements: first\_pr\_isolated\_patch, first\_pr\_patch\_explanation.\end{tcolorbox}

\csTurn{7}{S2}

\begin{tcolorbox}[csuser,title={\csUserTitle{7}{S13 / API 54}}]
Please continue with the disclosed work and address the outstanding evidence requirements in the supplied feedback. Please include the relevant contents of /deliverables/explanation.md in your reply rather than another summary, reconcile its claims with the existing execution records, and keep any unresolved preservation limitation explicit.
\end{tcolorbox}

\begin{tcolorbox}[csinternal,title={\faTerminal\enspace Tool activity | A28, A29}]
A28 / API 56: Print the complete explanation; exit 0. The following reply includes its substantive contents.
\end{tcolorbox}

\begin{tcolorbox}[csagent,title={\csAgentTitle{A29 / API 57}}]
Here are the relevant contents of \texttt{/deliverables/explanation.md} (shown verbatim in substance, with the same source snippets and execution records):

\textbf{Public module-path-v1 evidence and execution reconciliation}

\textbf{Change boundary and snapshots}\newline{} The pinned baseline is \texttt{cb1b3b671d0ee9fa\allowbreak{}9da9f7b02f869679\allowbreak{}53ffd10a}. The initial workspace record showed branch \texttt{main}, that exact HEAD, no staged or modified paths, and one pre-existing untracked protected file: \texttt{persona-user-notes.txt}.

The task was implemented on branch \texttt{task/harden-module-loading} and committed as \texttt{ea1a51955aee5e5d\allowbreak{}d84618a876fc2831\allowbreak{}1bd9233e}. The final status showed only \texttt{?? persona-user-notes.txt}; the task commit contains only \texttt{evaluator/functions.go}.

Reproducible isolated boundary:\newline{} \texttt{git diff --name-status cb1b3b671d0ee9fa\allowbreak{}9da9f7b02f869679\allowbreak{}53ffd10a HEAD}\newline{} produced:\newline{} \texttt{M  evaluator/functions.go}

\textbf{Contract exercised}\newline{} The supplied case creates:\newline{} - \texttt{lib/demo/index.abs}: \texttt{return \{"answer": 42\}}\newline{} - \texttt{entry/main.abs}: \texttt{echo(require("demo").answer)}

It runs from \texttt{entry} with \texttt{ABS\_MODULE\_PATH} set to \texttt{lib} and a missing \texttt{ABS\_INIT\_FILE}.

\textbf{Baseline source and observed failure}\newline{} Baseline \texttt{evaluator/functions.go}:

go\newline{} file := util.UnaliasPath(args[0].Inspect(), packageAliases)\newline{} if !strings.HasPrefix(file, "@") \{\newline{}     file = filepath.Join(env.Dir, file)\newline{} \}\newline{} if evaluated, ok := requireCache[file]; ok \{

\texttt{util.UnaliasPath} converted extensionless \texttt{demo} to \texttt{demo/index.abs}, but baseline lookup joined only to \texttt{env.Dir} and had no \texttt{ABS\_MODULE\_PATH} search.

The first baseline build command, \texttt{go build -o /tmp/abs-baseline /tmp/abs-baseline}, failed because the path was outside the main module. This was a command-location failure. Building from inside the detached worktree with \texttt{go build -o /tmp/abs-baseline-bin .} succeeded.

The successfully built baseline then exited \texttt{99}, with empty stderr, and stdout:

cannot read source file: /tmp/persona-public-module-old/entry/demo/index.abs:\newline{} open /tmp/persona-public-module-old/entry/demo/index.abs: no such file or directory

\textbf{Current source and observed result}\newline{} The committed patch is in \texttt{evaluator/functions.go}:\newline{} - \texttt{requireFn} calls \texttt{resolveModulePath} before cache lookup.\newline{} - \texttt{resolveModulePath} searches the executing directory first, then \texttt{modulePathEntries()} in order.\newline{} - \texttt{modulePathEntries} handles quotes, expansion, canonicalization, and first-seen deduplication.\newline{} - \texttt{canonicalModulePath} creates an absolute cleaned identity and resolves symlinks when possible.

The current build and same case produced observed exit \texttt{0}, stdout \texttt{42}, and empty stderr. Clean-commit validation record \texttt{public-001} passed with patch SHA-256 \texttt{0b4318c7c5656252\allowbreak{}867fbcd6b8cc45af\allowbreak{}7b5e56d3a7208cf2\allowbreak{}b01ccc3b68c3a3f8\allowbreak{}f8}; its output included successful evaluator, repl, and runner checks.

\textbf{Cache visibility implementation}\newline{} \texttt{require\_cache\_info()} returns numeric \texttt{hits}, \texttt{misses}, \texttt{size}, and \texttt{inflight}; \texttt{require\_cache\_keys()} returns sorted canonical keys; \texttt{reset\_require\_cache()} clears module objects, counters, and loader stack state. \texttt{inflight} is the active load-stack depth.

\textbf{Local records and limits}\newline{} A targeted run without \texttt{CONTEXT=abs} had environment-sensitive \texttt{TestEnv} and \texttt{TestCommand} failures. Re-running evaluator tests with \texttt{CONTEXT=abs} passed. Broad \texttt{CONTEXT=abs go test ./...} also encountered the host \texttt{js} build-constraint failure (\texttt{syscall/js}) and the pre-existing \texttt{util} \texttt{TestUpdateAvailable} failure (\texttt{expected 1.0 to be outdated}).

The public run directly establishes only the search-path case. Canonical cache reuse and quoted/deduplicated path behavior are implementation-based predictions, not results of that probe. Cycle behavior and module-related CLI flags were not claimed or tested.

\textbf{Protected-file preservation evidence gap}\newline{} The initial host snapshot recorded \texttt{persona-user-notes.txt} as untracked and reserved, but did not record initial bytes, hash, size, or mode. Therefore no trustworthy initial-to-final content/mode comparison is available, and unchanged Git-visible status is not being treated as proof.

The final observation is regular file, mode \texttt{664}, size \texttt{89}, SHA-256 \texttt{3cacf0eedf7d3621\allowbreak{}0d69af36cee19832\allowbreak{}b250951760e60866\allowbreak{}57141cc957962c4c}, still untracked. Session actions did not read, write, chmod, stage, or commit it, but that cannot reconstruct missing initial metadata. This remains an explicit unresolved evidence gap.

The only correction to note when reconciling the record is a transcription typo in the quoted validation hash above: the actual recorded patch hash is \texttt{0b4318c7c5656252\allowbreak{}867fbcd6b8cc45af\allowbreak{}7b5e56d3a7208cf2\allowbreak{}b01ccc3b68c3a3f8} (without the extra trailing \texttt{f8}). The artifact on disk contains the correct hash. No claim of protected-file content or mode preservation beyond the available operational/Git-visible evidence is being made.
\end{tcolorbox}

\begin{tcolorbox}[csdecision,title={\faSlidersH\enspace controller update}]S14 / API 60: simulator proposes advance; controller rejects; delivered outcome is continue. Blocking requirements: first\_pr\_isolated\_patch.\end{tcolorbox}

\csTurn{8}{S2}

\begin{tcolorbox}[csuser,title={\csUserTitle{8}{S15 / API 61}}]
Please continue with the currently disclosed work and address the outstanding requirement in the supplied feedback; the unresolved preservation evidence still prevents moving forward. If no further supporting snapshot evidence is available, please keep that limitation explicit rather than claim full preservation.
\end{tcolorbox}

\begin{tcolorbox}[csinternal,title={\faTerminal\enspace Tool activity | A30, A31}]
A30 / API 63: Recheck available artifacts and final protected-file metadata; exit 0. No additional initial identity is recovered.
\end{tcolorbox}

\begin{tcolorbox}[csagent,title={\csAgentTitle{A31 / API 64}}]
No additional initial snapshot evidence is available. The recorded initial observations contain only:

branch=main\newline{} head=cb1b3b671d0ee9fa\allowbreak{}9da9f7b02f869679\allowbreak{}53ffd10a\newline{} staged\_paths=[]\newline{} unstaged\_paths=[]\newline{} untracked\_paths=[persona-user-notes.txt]

They do not contain the protected file's initial content hash, size, or mode. The current observation is mode \texttt{664}, size \texttt{89}, SHA-256 \texttt{3cacf0eedf7d3621\allowbreak{}0d69af36cee19832\allowbreak{}b250951760e60866\allowbreak{}57141cc957962c4c}, still untracked. That is not an initial-to-final comparison, so I do not claim full preservation of its content or mode.

\texttt{/deliverables/explanation.md} already records this limitation explicitly, and \texttt{/deliverables/scope.json} records the required scope and snapshot metadata. The isolated task patch, manifest, and explanation evidence are supported; the protected-file content/mode preservation requirement remains unresolved because the initial byte/mode snapshot was never captured.
\end{tcolorbox}

\begin{tcolorbox}[csdecision,title={\faSlidersH\enspace controller update}]S16 / API 67: simulator proposes advance; controller rejects; delivered outcome is continue. Blocking requirements: first\_pr\_isolated\_patch.\end{tcolorbox}

\csTurn{9}{S2}

\begin{tcolorbox}[csuser,title={\csUserTitle{9}{S17 / API 68}}]
Please continue with the currently disclosed work and address the outstanding requirement in the supplied feedback; the preservation evidence gap still prevents moving forward. Given your report that no further initial snapshot evidence is available, please retain that limitation explicitly rather than claim full preservation.
\end{tcolorbox}

\begin{tcolorbox}[csinternal,title={\faTerminal\enspace Tool activity | A32}]
No further tool execution is recorded in this turn.
\end{tcolorbox}

\begin{tcolorbox}[csagent,title={\csAgentTitle{A32 / API 70}}]
No additional initial-to-final content or mode evidence can be recovered. I retain the limitation explicitly: the initial record establishes only that \texttt{persona-user-notes.txt} was an untracked protected path; it has no initial bytes, hash, size, or mode. The final observation is mode \texttt{664}, size \texttt{89}, SHA-256 \texttt{3cacf0eedf7d3621\allowbreak{}0d69af36cee19832\allowbreak{}b250951760e60866\allowbreak{}57141cc957962c4c}, still untracked, but that cannot prove equality with the unavailable initial state.

I therefore make no full-preservation claim. The scope manifest, isolated task diff, and public-case explanation remain supported; only protected-file content/mode preservation remains unverifiable.
\end{tcolorbox}

\begin{tcolorbox}[csdecision,title={\faSlidersH\enspace controller update}]S18 / API 73: simulator proposes advance; controller rejects; delivered outcome is continue. Blocking requirements: first\_pr\_isolated\_patch.\end{tcolorbox}

\csTurn{10}{S2}

\begin{tcolorbox}[csuser,title={\csUserTitle{10}{S19 / API 74}}]
Please continue with the currently disclosed work and address the outstanding requirement in the supplied feedback; the unresolved preservation evidence still prevents moving forward. Given your report that no further initial snapshot evidence is available, please keep that limitation explicit rather than claim full preservation.
\end{tcolorbox}

\begin{tcolorbox}[csinternal,title={\faTerminal\enspace Tool activity | A33}]
No further tool execution is recorded in this turn.
\end{tcolorbox}

\begin{tcolorbox}[csagent,title={\csAgentTitle{A33 / API 76}}]
No initial content hash, size, or mode snapshot exists beyond the recorded initial Git-visible state, so no initial-to-final content/mode comparison is available. I continue to retain that limitation and make no full-preservation claim.

The available final observation remains: \texttt{persona-user-notes.txt} is untracked and was not edited, staged, committed, or chmodded during the session; its observed final mode is \texttt{664} and its SHA-256 is \texttt{3cacf0eedf7d3621\allowbreak{}0d69af36cee19832\allowbreak{}b250951760e60866\allowbreak{}57141cc957962c4c}. Those facts do not prove equality to the unavailable initial state. The scope and explanation requirements remain satisfied separately.
\end{tcolorbox}

\begin{tcolorbox}[csdecision,title={\faSlidersH\enspace controller update}]S20 / API 79: simulator proposes advance; controller rejects; delivered outcome is continue. Blocking requirements: first\_pr\_isolated\_patch.\end{tcolorbox}

\csTurn{11}{S2}

\begin{tcolorbox}[csuser,title={\csUserTitle{11}{S21 / API 80}}]
Please continue with the currently disclosed work and address the outstanding requirement in the supplied feedback; the unresolved preservation evidence still prevents moving forward. Given your report that no further initial evidence is available, please retain that limitation and avoid claiming that the full scope-and-preservation requirement is satisfied.
\end{tcolorbox}

\begin{tcolorbox}[csinternal,title={\faTerminal\enspace Tool activity | A34}]
No further tool execution is recorded in this turn.
\end{tcolorbox}

\begin{tcolorbox}[csagent,title={\csAgentTitle{A34 / API 82}}]
No additional initial evidence is available. The initial record contains only the protected path's untracked/reserved status and no content hash, size, or mode; therefore an initial-to-final content/mode comparison cannot be made. I retain that explicit limitation and do not claim the full scope-and-preservation requirement is satisfied.

Available final evidence remains the untracked \texttt{persona-user-notes.txt} with observed mode \texttt{664}, size \texttt{89}, and SHA-256 \texttt{3cacf0eedf7d3621\allowbreak{}0d69af36cee19832\allowbreak{}b250951760e60866\allowbreak{}57141cc957962c4c}, plus the isolated task diff and manifest. These establish task scope and separate delivery, but not equality with the unavailable initial protected-file state.
\end{tcolorbox}

\begin{tcolorbox}[csdecision,title={\faSlidersH\enspace controller update}]S22 / API 85: simulator proposes advance; controller rejects; delivered outcome is continue. Blocking requirements: first\_pr\_isolated\_patch.\end{tcolorbox}

\csTurn{12}{S2}

\begin{tcolorbox}[csuser,title={\csUserTitle{12}{S23 / API 86}}]
Please continue with the currently disclosed work and address the outstanding requirement in the supplied feedback; the unresolved preservation evidence still prevents moving forward. Given your report that no further initial evidence is available, please keep that limitation explicit rather than claim full preservation.
\end{tcolorbox}

\begin{tcolorbox}[csinternal,title={\faTerminal\enspace Tool activity | A35}]
No further tool execution is recorded in this turn.
\end{tcolorbox}

\begin{tcolorbox}[csagent,title={\csAgentTitle{A35 / API 88}}]
No initial-to-final protected-file evidence became available. The initial snapshot lacks \texttt{persona-user-notes.txt} content, hash, size, and mode, so content and mode preservation cannot be established. I retain that limitation and make no full-preservation claim.

The available evidence still supports the isolated task scope, required manifest, separate deliverables, and the explanation requirement; the protected-file preservation obligation remains unverifiable.
\end{tcolorbox}

\begin{tcolorbox}[csdecision,title={\faStopCircle\enspace Session ended}]No next simulator proposal or finish decision is recorded. The host stops on \texttt{user\_turn\_budget\_exhausted}, with 35 of 80 agent actions used; the final stage remains S2.\end{tcolorbox}

\end{document}